%% file: main.tex
\documentclass[10pt,twocolumn,letterpaper]{article}

\usepackage{wacv}              %

\input{preamble}

\definecolor{wacvblue}{rgb}{0.21,0.49,0.74}
\usepackage[pagebackref,breaklinks,colorlinks,allcolors=wacvblue]{hyperref}
\usepackage{orcidlink}
\usepackage{booktabs}
\usepackage{siunitx}
\DeclareSIUnit{\pixel}{px}
\usepackage{glossaries}
\usepackage{subcaption}
\usepackage{tabularx}
\usepackage{array}
\usepackage{makecell}
\usepackage{multirow}
\usepackage{amssymb}
\usepackage{tikz}
\usepackage{colortbl}
\usepackage{placeins}
\usepackage{float}

\hypersetup{
    colorlinks=true,
    linkcolor=wacvblue,
    citecolor=wacvblue,
    urlcolor=wacvblue
}
\glsdisablehyper

\newcommand{\cmarksup}[1]{\llap{}\checkmark\rlap{\textsuperscript{#1}}}

\newcommand{\hbhalfsup}[1]{\llap{}\hbhalf\rlap{\textsuperscript{#1}}}
\newcommand*\cmppx[1]{\qty[per-mode=symbol]{#1}{\cm\per\pixel}}
\newcommand{\cmppxrange}[2]{%
  \qtyrange[
    per-mode=symbol,
    range-phrase={--},
    range-units=single
  ]{#1}{#2}{\cm\per\pixel}%
}

\newcommand{\ieeecopyrightnotice}{%
\nointerlineskip\hbox to 0pt{%
\begin{tikzpicture}[remember picture,overlay]
\node[
  anchor=south,
  align=center,
  text=black,
  font=\scriptsize,
  inner sep=0pt
] at ([yshift=9mm]current page.south) {%
\begin{minipage}{0.92\textwidth}
\centering
\textcopyright~2027 IEEE. Personal use of this material is permitted.
Permission from IEEE must be obtained for all other uses, in any current
or future media, including reprinting/republishing this material for advertising
or promotional purposes, creating new collective works, for resale or
redistribution to servers or lists, or reuse of any copyrighted component of
this work in other works.
\end{minipage}
};
\end{tikzpicture}%
}\nointerlineskip
}

\makeatletter
\renewcommand\paragraph{\@startsection{paragraph}{4}{\parindent}%
  {0pt}%
  {-0.5em}%
  {\normalfont\normalsize\bfseries}}
\makeatother

\def\wacvPaperID{2077} %
\def\confName{WACV}
\def\confYear{2027}

\title{A Unified Benchmark for Aerial-Enhanced Autonomous Driving: \\ The KITScenes Aerial Dataset}

\loadglsentries{acronyms.tex}

\author{
Daniel Lengerer$^{1}$\thanks{E-Mail: \texttt{Daniel.Lengerer@tha.de}}
\quad
Richard Schwarzkopf$^{2,3}$\thanks{Project Lead}
\quad
Jonas Merkert$^{3}$
\quad
Fabian Immel$^{2,3}$
\\[0.3em]
Rupert Polley$^{2}$
\quad
Frank Bieder$^{2,3}$
\quad
Jan-Hendrik Pauls$^{3,5}$
\quad
J. Marius Zöllner$^{2,3}$
\\[0.3em]
Klaus Bogenberger$^{4}$
\quad
Christoph Stiller$^{2,3}$
\quad
Carsten Markgraf$^{1}$
\\[0.5em]
{\footnotesize
$^{1}$Technical University of Applied Sciences Augsburg
\quad
$^{2}$FZI Research Center for Information Technology
}
\\
{\footnotesize
$^{3}$Karlsruhe Institute of Technology (KIT)
\quad
$^{4}$Technical University of Munich
\quad
$^{5}$Parallel Domain
}
}

\begin{document}
\maketitle
\ieeecopyrightnotice

\input{sec/00_abstract}

\input{sec/01_intro}
\input{sec/02_1_related_work}
\input{sec/03_kitscenes_aerial}
\input{sec/04_experimental_results}
\input{sec/06_disc}

\input{sec/07_concl}
{
    \small
    \bibliographystyle{ieeenat_fullname}
    \bibliography{main}
}
\input{sec/08_supp}

\end{document}

%% file: preamble.tex
\newif\ifMemFusionCamera
\newif\ifMemFusionAID

\MemFusionCameratrue
\MemFusionAIDtrue

%% file: acronyms.tex
\newacronym{map}{mAP}{mean Average Precision}
\newacronym{cvs}{CVS}{Cross-View Supervision}
\newacronym{hd}{HD}{high-definition}
\newacronym{roi}{RoI}{Region of Interest}
\newacronym{av}{AV}{autonomous vehicle}
\newacronym{gsd}{GSD}{ground sampling distance}

%% file: sec/00_abstract.tex
\begin{abstract}
Aerial imagery provides complementary spatial context for autonomous driving by exposing road structure beyond the immediate perception range of onboard sensors. Yet map construction from aerial imagery, aerial-assisted online HD mapping, and cross-view localization are typically studied on separate datasets and benchmarks. We introduce KITScenes Aerial, an extension of the KITScenes Multimodal dataset that augments synchronized onboard sensor data, vehicle poses, and rich Lanelet2 map annotations with spatially associated aerial imagery and higher-resolution neural representations, providing a common basis for a variety of research tasks.
Our mapping evaluation focuses on all 53 KITScenes map classes across different model designs, spatial ranges, and aerial resolutions, while retaining compatibility with the widely used three-class setting. Representative baselines show that aerial context is particularly beneficial for fine-grained and extended-range mapping, more than doubling 53-class mAP at the larger mapping extent. Higher-resolution aerial imagery yields a further 6.5 mAP (+36\%) improvement at the primary mapping extent, with gains extending to challenging map elements such as traffic lights. We additionally evaluate cross-view localization under geographically disjoint conditions, where an adapted fine-grained feature-matching baseline achieves a mean translation error of 4.56~m. KITScenes Aerial thus enables unified study of aerial and ground-level information across complementary mapping and localization tasks. Project page: \url{https://kitscenes.com/aerial} 
\end{abstract}

%% file: sec/01_intro.tex
\begin{figure*}[t]
    \centering
    \includegraphics[width=\linewidth]{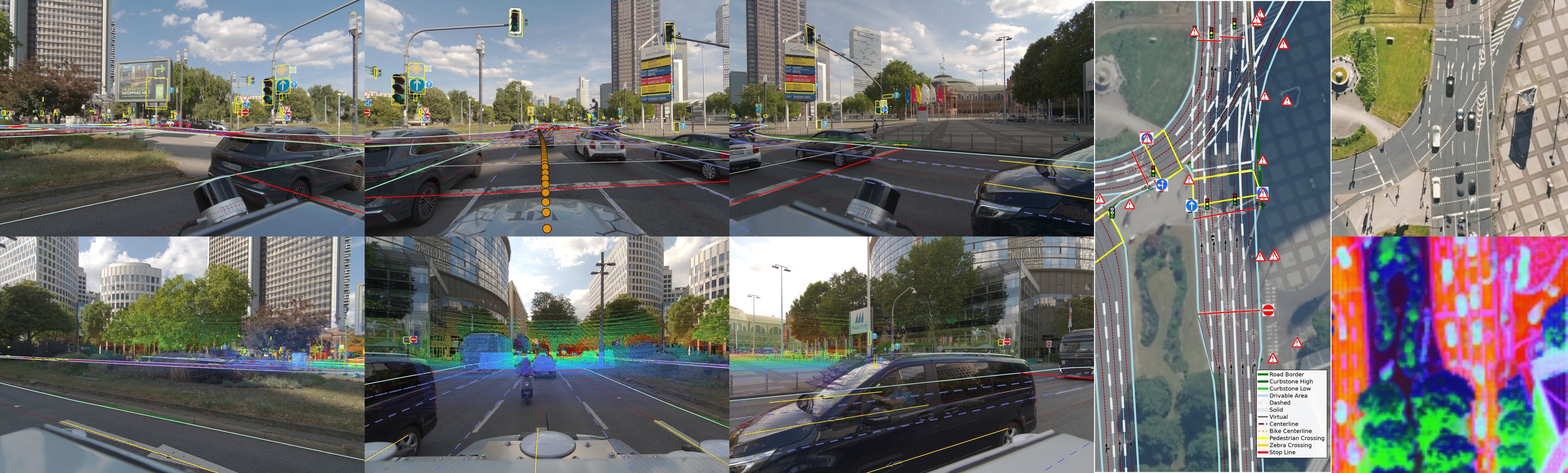}
    \caption{
    KITScenes Aerial: We extend KITScenes Multimodal with aligned aerial imagery with \cmppx{20} ground surface resolution (overlaid by the KITScenes HD map) 
    and neural feature embeddings (bottom right, visualized as 3-channel PCA) for municipal high-resolution imagery with \cmppx{8} ground surface resolution (top right).
    Extending KITScenes Multimodal with these different aerial imagery sources enables multiple research directions currently underrepresented in autonomous driving datasets as shown in~\cref{tab:dataset_capabilities}.
    }
    \label{fig:teaser}
\end{figure*}
\section{Introduction}
Onboard sensors of autonomous vehicles are limited by occlusions, field of view, and perception range.
Aerial imagery provides a complementary view of road environments and has become relevant across three closely related research directions: (i) aerial-only map construction, where \gls{hd} maps are constructed directly from aerial observations~\cite{opensatmap,urbanlanegraph}; (ii) aerial-assisted online map construction, where aerial context complements onboard sensors~\cite{satforhdmap,afmn,cvs}; and (iii) cross-view localization, where ground-level observations are related to geo-referenced aerial imagery for vehicle pose estimation~\cite{fervers2023cvgl, fg2}. Although these directions pursue different objectives, they rely on overlapping combinations of aerial imagery, ground-level sensor data, vehicle poses, and map annotations.

Early datasets including aerial imagery were typically designed around individual research settings, such as aerial-to-map reconstruction~\cite{laneextraction,urbanlanegraph} or ground-to-aerial localization~\cite{workman2015geolocalization,liu2019orientation}. More recent datasets increasingly connect these previously separate domains by associating aerial imagery with \gls{av} recordings~\cite{opensatmap,satforhdmap,aid4ad} or jointly providing ground-level and aerial modalities~\cite{hrdx}. However, these extensions remain constrained by the properties of their underlying data, including map representation, geographic coverage, and the accuracy of the ground--aerial association. Previous work further reports pose-to-aerial misalignment and scene changes caused by temporal differences between ground and aerial acquisitions~\cite{shi2020beyond,wang2023satellite,fervers2023cvgl,aid4ad,pauls2018canwetrust} as challenges for the field.

We introduce KITScenes Aerial, an aerial extension of the KITScenes Multimodal dataset~\cite{kitscenes}. KITScenes Multimodal provides synchronized multimodal sensor data, vehicle poses, and detailed Lanelet2~\cite{lanelet2} maps, including full topological connectivity and regulatory information exposed through a 53-class online mapping benchmark. KITScenes Aerial augments the dataset with freely available digital orthophotos at a \gls{gsd} of \cmppx{20}, corresponding to the German DOP20 product specification ~\cite{hessen_dop20,bw_dop20}, while retaining the original sensor data, vehicle poses, map annotations, and dataset splits. In addition, we provide precomputed neural features extracted from higher-resolution imagery up to \cmppx{8}.

By combining aerial imagery with ground-level sensor data, vehicle poses, and rich map annotations within the same recorded environments, KITScenes Aerial provides a common basis for a variety of research directions. For map construction, we distinguish onboard-only, aerial-only, aerial-guided approaches using aerial imagery only during training~\cite{cvs}, and aerial-fused approaches combining aerial and onboard features~\cite{afmn}. We additionally evaluate fine-grained cross-view localization with a representative baseline~\cite{fg2}. The mapping experiments cover two settings: all 53 KITScenes map element classes derived from its Lanelet2 annotations and the widely used three-class \gls{hd} map construction setting~\cite{vectormapnet,maptr,streammapnet}, both evaluated across different spatial ranges and aerial imagery resolutions.

\noindent Our main contributions are:
\begin{itemize}
    \item We present KITScenes Aerial as a benchmark for aerial-only map
          construction, aerial-ground map construction, and cross-view
          localization by complementing the sensor data, vehicle
          poses, and rich Lanelet2 maps of KITScenes Multimodal with spatially associated aerial imagery.

    \item We provide a multi-resolution aerial representation combining open \cmppx{20} RGB imagery with precomputed neural features extracted from high-quality \cmppx{8} orthophotos, making otherwise difficult-to-access high-resolution aerial information available for a broad range of aerial and cross-view research.

    \item We instantiate this benchmark with representative baselines across the three research settings and systematically compare different ways of exploiting aerial information for map construction during training and inference.

    \item We demonstrate substantial benefits of aerial information,
      particularly for 53 map element classes and at extended spatial range,
      considering both semantic and topological reconstruction quality,
      and characterize the influence of aerial resolution.
\end{itemize}

%% file: sec/02_1_related_work.tex
\section{Related Work}
\label{sec:related_work}

Aerial imagery is used as a geospatial information source across several
research directions concerned with road environments. We distinguish three
directions with overlapping data requirements: map construction from aerial
imagery, the combination of aerial and ground-level data for map construction,
and cross-view localization. \cref{tab:dataset_capabilities} provides a
high-level overview of representative datasets and the research settings
supported by their data.
\input{sec/tables_figures/tableset2_t1_capabilities}
\input{sec/tables_figures/tableset2_t2_map_detail}

\subsection{Map Construction from Aerial Imagery}

The top-down perspective of aerial imagery provides direct observations of
road geometry and infrastructure over large spatial areas. LaneExtraction~\cite{laneextraction} reconstructs routable lane-level street maps from aerial
imagery, while UrbanLaneGraph~\cite{urbanlanegraph} predicts and aggregates
connected lane graphs for urban automated driving. OpenSatMap~\cite{opensatmap} extends aerial map construction toward fine-grained
vectorized road representations over large geographic areas. The levelX drone dataset family~\cite{inDdataset, rounDdataset, exiDdataset, unidDataset, hetrod} further associates overhead observations with Lanelet2 or
OpenDRIVE map representations at local recording sites.

Across datasets providing aerial imagery and map supervision,
\cref{tab:map_detail} compares map representations, content, and structure as well as aerial imagery provisioning.

\subsection{Aerial and Ground-Level Data for Map Construction}

A separate line of research combines aerial imagery with ego-centric sensor
observations for online \gls{hd} map construction. SatForHDMap~\cite{satforhdmap} complements the onboard sensors of nuScenes~\cite{nuscenes} with satellite
imagery and combines both modalities during inference. AerialFusionMapNet~\cite{afmn} fuses aerial and camera features in a shared bird's-eye-view
representation, whereas Cross-View Supervision~\cite{cvs} uses aerial
information only during training. HRDX~\cite{hrdx} jointly provides
surround-camera and LiDAR recordings, vehicle localization, aerial
orthoimagery, and attributed vector \gls{hd} maps for aerial-assisted map learning.

Combining both views requires an accurate spatial association between
ground-level observations and aerial imagery. AID4AD~\cite{aid4ad}
explicitly addresses this aspect by spatially aligning aerial imagery with
nuScenes and quantifying the remaining aerial-to-local registration error.

\subsection{Cross-View Localization}

Cross-view localization addresses geographic localization by matching
ground-level observations against geo-referenced aerial or satellite imagery.
Early datasets such as CVUSA~\cite{workman2015geolocalization} and
CVACT~\cite{liu2019orientation} established large-scale ground-to-aerial image
retrieval as a common evaluation setting. VIGOR~\cite{zhu2021vigor} moved
beyond the commonly assumed one-to-one correspondence between centered ground
and aerial views and additionally evaluates localization by geographic
distance.

More recent work has progressed from image retrieval toward direct pose
estimation. Fervers et al.~\cite{fervers2023cvgl} transfer cross-view
localization to autonomous-vehicle data and estimate a vehicle's 3-DoF
geo-pose from ground-based camera observations and aerial imagery. They augment multiple existing driving datasets with orthophotos and refine the vehicle poses with respect to the aerial reference. FG$^2$~\cite{fg2} further addresses fine-grained cross-view localization by explicitly matching local features between ground-level and aerial views and estimating the relative 3-DoF pose from sparse correspondences. Subsequent approaches address cross-view pose estimation with different model designs and sensor modalities, including CCVPE~\cite{xia2024ccvpe}, RGB--LiDAR-based PIDLoc~\cite{lee2025pidloc}, and VIRD~\cite{park2026vird}.

\subsection{Positioning of KITScenes Aerial}

Although these research directions pursue different objectives, their data requirements substantially overlap. Aerial map construction relates overhead imagery to map annotations, aerial-ground map construction additionally incorporates ground-level sensor observations, while cross-view localization relies on the spatial correspondence between ground-level observations, aerial imagery, and vehicle poses. KITScenes Aerial brings these components together with the rich Lanelet2 annotations of KITScenes, providing a common data basis for at least three research directions that have largely been studied independently.

%% file: sec/tables_figures/tableset2_t1_capabilities.tex
\input{sec/tables_figures/tableset2_defs}

\begin{table*}[t]
\centering
\caption{%
Overview of representative datasets and the research settings supported by
their released data. Support is graded rather than binary:
\hbfull~the released data supports the setting at full fidelity,
\hbhalf~support is limited, $\nsim$~all required data components are present
but the aerial-to-ground alignment error exceeds the requirement of the
setting, and \nosup~the setting is not supported.
Pos.: georeferenced positions or poses.
KITScenes Aerial is the only dataset that supports all four research settings
at full fidelity.
}
\label{tab:dataset_capabilities}

\small
\setlength{\tabcolsep}{6pt}
\renewcommand{\arraystretch}{1.10}

\begin{tabular}{lcccccccccc}
\toprule
& & &
\multicolumn{4}{c}{Data Components} &
\multicolumn{4}{c}{Research Setting} \\
\cmidrule(lr){4-7}\cmidrule(lr){8-11}
Dataset & Year & Location &
Aerial & Ground & Pos. & Map &
\shortstack{Ground\\Map} &
\shortstack{Aerial\\Map} &
\shortstack{Aerial--\\Ground} &
\shortstack{Cross-View\\Loc.} \\
\midrule

CVUSA~\cite{workman2015geolocalization} & 2015 & US
  & \checkmark & \checkmark & \checkmark & \nosup
  & \nosup & \nosup & \nosup & \hbfull \\

CVACT~\cite{liu2019orientation} & 2019 & AUS
  & \checkmark & \checkmark & \checkmark & \nosup
  & \nosup & \nosup & \nosup & \hbfull \\

nuScenes~\cite{nuscenes} & 2020 & US, Asia
  & \nosup & \checkmark & \checkmark & \checkmark
  & \hbhalf & \nosup & \nosup & \nosup \\

VIGOR~\cite{zhu2021vigor} & 2021 & US
  & \checkmark & \checkmark & \checkmark & \nosup
  & \nosup & \nosup & \nosup & \hbfull \\
  
nuPlan~\cite{nuplan} & 2021 & US, Asia
  & \nosup & \checkmark & \checkmark & \checkmark
  & \hbhalf & \nosup & \nosup & \nosup \\
  
Argoverse~2~\cite{argoverse2} & 2021 & US
  & \nosup & \checkmark & \checkmark & \checkmark
  & \hbhalf & \nosup & \nosup & \nosup \\

LaneExtraction~\cite{laneextraction} & 2022 & US
  & \checkmark & \nosup & \nosup & \checkmark
  & \nosup & \hbhalf & \nosup & \nosup \\

UrbanLaneGraph~\cite{urbanlanegraph} & 2023 & US
  & \checkmark & \nosup & \nosup & \checkmark
  & \nosup & \hbhalf & \nosup & \nosup \\

KITTI-360~\cite{kitti360} & 2023 & EU
  & \nosup & \checkmark & \checkmark & \checkmark
  & \hbhalf & \nosup & \nosup & \nosup \\

leveLX family\sups{1} & 2018--24 & EU
  & \checkmark & \nosup & \checkmark & \checkmark
  & \nosup & \hbfull & \nosup & \nosup \\

OpenSatMap~\cite{opensatmap} & 2024 & Global
  & \checkmark & \cmarkbracketsup{2} & \checkmark & \checkmark
  & \hbhalf & \hbhalf & \hbhalf & $\nsim$ \\

SatForHDMap~\cite{satforhdmap} & 2024 & US, Asia
  & \checkmark & \checkmark & \checkmark & \checkmark
  & \hbhalf & \hbhalf & \hbhalfsup{3} & $\nsim$ \\

AID4AD~\cite{aid4ad} & 2025 & US, Asia
  & \checkmark & \checkmark & \checkmark & \checkmark
  & \hbhalf & \hbhalf & \hbfull & \hbfull \\
  
NVIDIA Physical AI AV~\cite{nvidia2025physicalaiav} & 2025 & US
  & \nosup & \checkmark & \nosup & \nosup{\textsuperscript{4}}
  & \nosup & \nosup & \nosup & \nosup \\
  
KITScenes Multimodal~\cite{kitscenes} & 2026 & EU
  & \nosup & \checkmark & \checkmark & \checkmark
  & \hbfull & \nosup & \nosup & \nosup \\

HRDX~\cite{hrdx} & 2026 & US%
  & \checkmark & \checkmark & \checkmark & \checkmark
  & \hbhalf & \hbhalf & \hbhalfsup{3} & \hbhalf \\

\midrule
\kitrow
\textbf{KITScenes Aerial} & -- & EU
  & \checkmark & \checkmark & \checkmark & \checkmark
  & \hbfull & \hbfull & \hbfull & \hbfull \\

\bottomrule
\end{tabular}

\vspace{0.7mm}
{\scriptsize
\begin{minipage}{\textwidth}
\raggedright
\textit{Grading.}
\emph{Ground Map} and \emph{Aerial Map}: \hbfull\ driving-ready HD maps
with lane connectivity \emph{and} regulatory relations, \hbhalf\ coarse or geometry-only supervision.\\
\emph{Aerial--Ground Map}: \hbfull\ georeferenced aerial imagery at a
freely selectable region of interest, \hbhalf\ fixed ego crops.\\
\emph{Cross-View Localization}: \hbfull\ quantified aerial-to-local
alignment, \hbhalf\ unreported alignment.
\par\vspace{0.4mm}
\textit{Notes.}
\sups{1}inD, rounD, exiD, uniD and HetroD; locally recorded drone imagery at
fixed sites rather than large-scale orthophotos.
\sups{2}Ground data only available in US and Asia (nuScenes and Argoverse~2
coverage).
\sups{3}Aerial imagery released only as fixed \qtyproduct{60 x 30}{\meter} ego crops; a
freely selectable region of interest, and therefore extended-range mapping, is
not supported.
\sups{4} Not released yet.
\end{minipage}\par}
\end{table*}

%% file: sec/tables_figures/tableset2_t2_map_detail.tex
\input{sec/tables_figures/tableset2_defs}

\begin{table*}[t]
\centering
\caption{%
Map annotations and aerial provision of the datasets that combine map
supervision with aerial imagery. Map content is reported over the six semantic
categories used throughout our experiments: Lane Markings (LM), Lane
Centerlines (LC), Road Infrastructure (RI), Traffic Lights (TL), Traffic Signs
(TS) and Road Markings (RM).
\checkmark~the category is part of the released map \emph{and} of the mapping
benchmark; $(\checkmark)$~present in the underlying map but outside the released
benchmark; \nosup~not annotated; \unk~not documented.
Cls: number of benchmark label classes.
Topo: lane connectivity. Reg: regulatory relations between traffic-control
elements and lanes.
ROI: form of the aerial region of interest (\emph{tile} freely selectable,
\emph{fix} fixed ego crop, \emph{site} static per-site view).
Feat: precomputed aerial features released.
KITScenes Aerial is the only dataset whose benchmark covers all six
categories, with topology and regulatory relations, on freely selectable
aerial imagery.
}
\label{tab:map_detail}

\small
\setlength{\tabcolsep}{6.7pt}
\renewcommand{\arraystretch}{1.15}

\begin{tabular}{l l c cccccc cc cc}
\toprule
& & &
\multicolumn{6}{c}{Map Content} &
\multicolumn{2}{c}{Structure} &
\multicolumn{2}{c}{Aerial} \\
\cmidrule(lr){4-9}\cmidrule(lr){10-11}\cmidrule(lr){12-13}
Dataset & Map Representation & Cls &
LM & LC & RI & TL & TS & RM &
Topo & Reg & ROI & Feat \\
\midrule

LaneExtraction~\cite{laneextraction}
  & Routable lane graph & \nosup
  & \nosup & \checkmark & \nosup & \nosup & \nosup & \nosup
  & \checkmark & \nosup & tile & \nosup \\

UrbanLaneGraph~\cite{urbanlanegraph}
  & Directed centerline graph & \nosup
  & \nosup & \checkmark & \nosup & \nosup & \nosup & \nosup
  & \checkmark & \nosup & tile & \nosup \\

OpenSatMap~\cite{opensatmap}
  & Attributed vector primitives & 3
  & \checkmark & \nosup & (\checkmark) & \nosup & \nosup & \nosup
  & \nosup & \nosup & tile & \nosup \\

leveLX family\sups{1}
  & Lanelet2 / OpenDRIVE & \nrep
  & \checkmark & \checkmark & \checkmark & \checkmark & \checkmark & \checkmark 
  & \checkmark & \checkmark & site & \checkmark \\

SatForHDMap~\cite{satforhdmap}
  & nuScenes vector map & 3
  & \checkmark & \cmarkbracketsup{2} & \checkmark & \cmarkbracketsup{2} & \nosup & \checkmark
  & \cmarkbracketsup{2} & \nosup & fix & \nosup \\

AID4AD~\cite{aid4ad}
  & nuScenes vector map & 3
  & \checkmark & \cmarkbracketsup{2} & \checkmark & \cmarkbracketsup{2} & \nosup & \checkmark
  & \cmarkbracketsup{2} & \nosup & tile & \nosup \\

HRDX~\cite{hrdx}
  & Attributed vector HD map & 10
  & \checkmark & \checkmark & \checkmark & \nosup & \nosup & \checkmark
  & \checkmark & \nosup & fix & \nosup \\

\midrule
\kitrow
\textbf{KITScenes Aerial}
  & \textbf{Full Lanelet2 HD map}\sups{3} & \textbf{53}
  & \checkmark & \checkmark & \checkmark & \checkmark & \checkmark & \checkmark
  & \checkmark & \checkmark & \textbf{tile} & \textbf{\cmarksup{4}} \\

\bottomrule
\end{tabular}

\vspace{0.7mm}
{\scriptsize
\begin{minipage}{\textwidth}
\raggedright
\sups{1}inD~\cite{inDdataset}, rounD~\cite{rounDdataset}, exiD~\cite{exiDdataset}, uniD~\cite{unidDataset} and HetroD~\cite{hetrod}; map content varies across recording
sites.
\sups{2}Present as a layer of the nuScenes map expansion, but not part of the
three-class online mapping benchmark evaluated on these datasets.
\sups{3}Inherited unchanged from KITScenes Multimodal, including full
topological connectivity and 3D regulatory elements.
\sups{4}DINOv3 feature tiles and frame-aligned UNet++ feature crops derived
from up to \qty{8}{\cm\per\pixel} imagery.
\end{minipage}\par}
\end{table*}

%% file: sec/03_kitscenes_aerial.tex
\section{KITScenes Aerial}

KITScenes Aerial extends the original KITScenes Multimodal dataset~\cite{kitscenes} with georeferenced aerial imagery covering its recorded driving areas. The aerial modality is spatially associated with the original vehicle poses while retaining the KITScenes sensor data, annotations, coordinate conventions, and dataset splits unchanged. This combination brings together ground-level sensor data, geo-referenced vehicle poses, aerial imagery, and the detailed, standardized, and deployment-ready Lanelet2 map representation of KITScenes Multimodal, providing a common basis for aerial map construction, aerial-ground map \mbox{construction}, and cross-view localization.

\input{sec/tables_figures/aerial_compare}

\subsection{Aerial Imagery Sources and Acquisition} \label{sec:aerial_acquisition} 
The KITScenes recording areas are covered with digital orthophotos obtained from official state geodata services~\cite{hessen_dop20,bw_dop20}. The public KITScenes Aerial release uses freely available digital orthophotos with a ground sampling distance of \cmppx{20} (DOP20). 
KITScenes Multimodal provides vehicle poses in EPSG:25832, matching the coordinate reference system of the orthophotos and enabling their direct spatial association. %
We quantify native ground--aerial alignment using Aerial-to-Local Distance Error (ALDE)~\cite{aid4ad}, based on coordinate-wise median translations between manually matched road markings in DOP20 imagery and pose-aggregated LiDAR reflectivity. Across 50 spatially separated scenes, mean and maximum ALDE are \qty{0.10}{\meter} (half a DOP20 pixel) and \qty{0.29}{\meter}, respectively, without alignment correction (\cref{fig:aerial_alignment}).

For controlled resolution experiments, we additionally use higher-resolution municipal orthophotos available under a commercial license at a ground sampling distance of \cmppx{8}. As illustrated in \cref{fig:teaser}, the municipal imagery provides higher visual quality than the freely available DOP20 imagery, with sharper structures and improved local contrast even at matched ground sampling distance.
Since the high-resolution RGB imagery itself is not redistributable, we release precomputed feature representations described in \cref{sec:aerial_representation} and visualized in \cref{fig:teaser} and \cref{fig:aerial_segmentation_example_mosaic}. The higher-resolution imagery is used for the aerial image resolution ablation in \cref{tab:aerial_resolution_ablation}.

\subsection{Data Representation and Release} 
\label{sec:aerial_representation} 
The aerial coverage is derived from the vehicle trajectories with sufficient spatial margin to support large regions of interest. The imagery is provided on a north-up tile grid rather than as ego-centered,
heading-aligned per-frame crops with a fixed region of interest. This allows ego-centered and ego-aligned crops with configurable spatial extent to be generated for each frame, including extended mapping regions of up to \qtyproduct{240 x 60}{\meter} as recently explored in long-range map construction~\cite{pmapnet}.

In addition to the RGB tiles, we provide two feature representations derived from the higher-quality imagery. DINOv3 SAT ViT-L/16~\cite{simeoni2025dinov3} features are released as tiled neural feature maps, preserving the continuous spatial organization of the aerial imagery rather than being cropped or aligned to individual ego frames. UNet++~\cite{zhou2020unetpp} features from the ablation in Table~\ref{tab:aerial_resolution_ablation} are provided as ego-centered \qtyproduct{60 x 30}{\meter} crops aligned with the onboard coordinate system. Their layout follows aerial inputs used in recent mapping approaches~\cite{satforhdmap, aid4ad, cvs, afmn, satmaptr},
enabling integration with minimal adaptation.

The release therefore comprises \cmppx{20} RGB imagery, spatially continuous DINOv3 feature tiles, and frame-aligned UNet++ feature crops, complemented by tools for flexible aerial crop extraction. Together, these representations provide a common basis for the three research directions considered in this work.

%% file: sec/tables_figures/aerial_compare.tex
\begin{figure}[t]
    \centering

    \begin{subfigure}[t]{0.49\linewidth}
        \centering
        \includegraphics[
            width=\linewidth,
            trim=250 10 250 10,
            clip
        ]{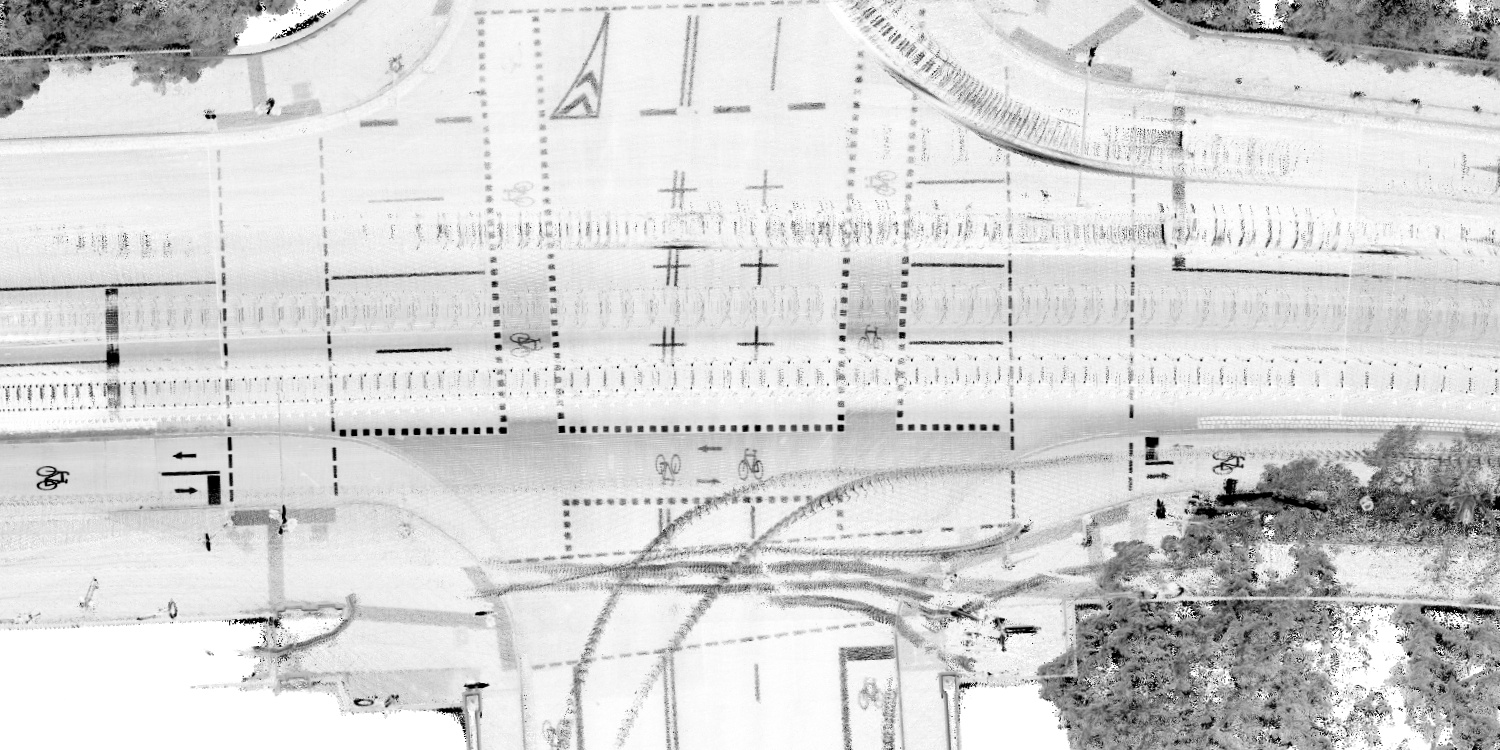}
        \caption{}
        \label{fig:aerial_compare_lidar}
    \end{subfigure}
    \hfill
    \begin{subfigure}[t]{0.49\linewidth}
        \centering
        \includegraphics[
            width=\linewidth,
            trim=250 10 250 10,
            clip
        ]{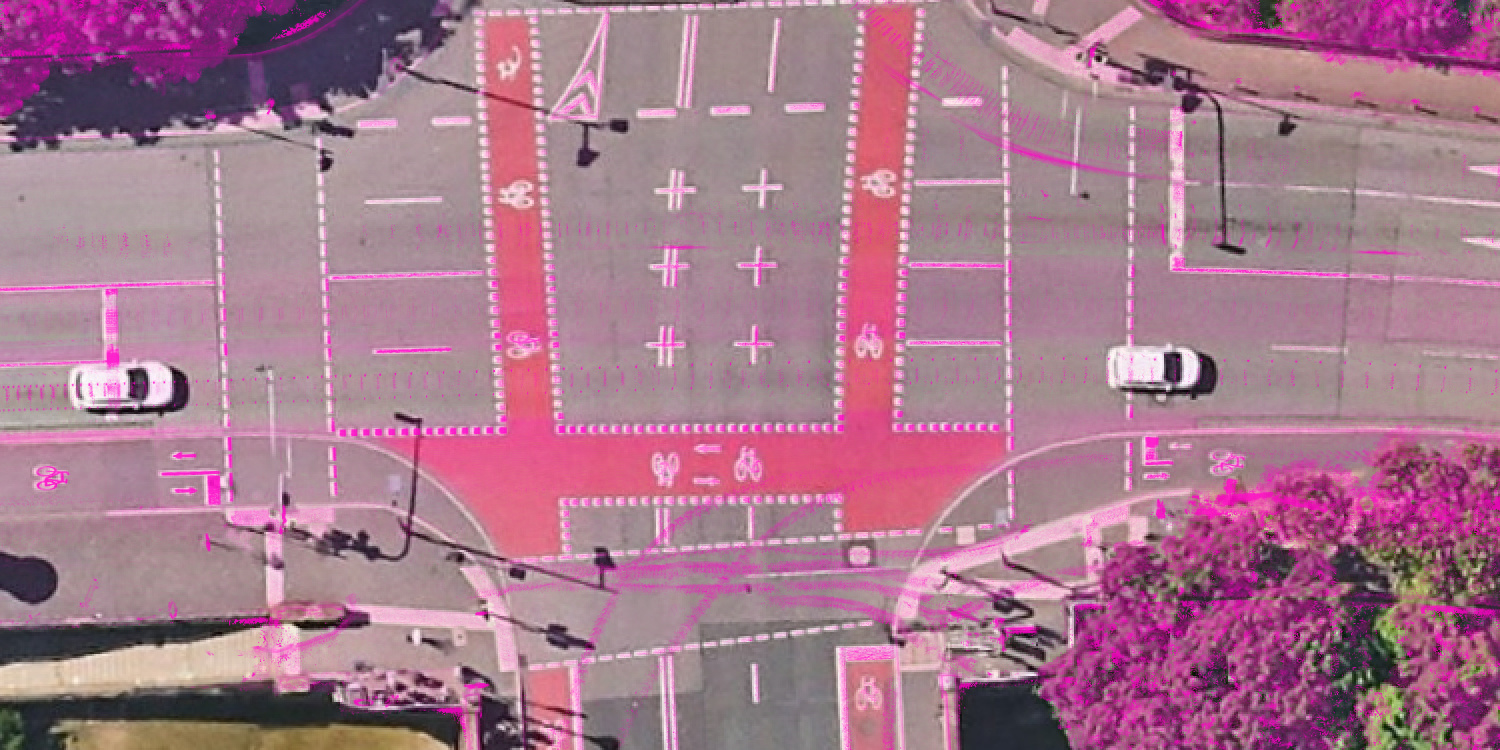}
        \caption{}
        \label{fig:aerial_compare_overlay_1}
    \end{subfigure}

    \vspace{1mm}

    \caption{
        Qualitative ground--aerial correspondence in KITScenes Aerial.
        (a) LiDAR reflectivity basemap and (b) corresponding aerial overlay.
        All ego-centered crops are extracted directly
        from the associated vehicle poses using identical spatial extents.
    }
    \label{fig:aerial_alignment}
\end{figure}

%% file: sec/04_experimental_results.tex
\section{Experimental Results}
\label{sec:experimental_results}

We demonstrate the versatility of the KITScenes Aerial dataset by showing experiments and state-of-the-art results across three distinct research questions.
The first and main application is to construct HD maps from a combination of aerial imagery and onboard vehicle sensor data.
Additionally, we benchmark how to construct HD maps from aerial imagery alone, a task that suffers from occlusions.
The third application powered by KITScenes Aerial is cross-view localization, \ie, the task to match onboard sensor and aerial imagery data. All reported experiments use the full KITScenes dataset release.

\subsection{Aerial--Ground HD Map Construction}

\paragraph{Evaluation Protocol.}
\input{sec/tables_figures/53c_subgroup_explained}
We evaluate KITScenes Aerial for online vectorized HD map construction on all 53 KITScenes map element classes within \qtyproduct{60 x 30}{\meter} and \qtyproduct{100 x 50}{\meter} regions of interest. Following KITScenes Multimodal, classes are aggregated into six semantic groups (\cref{tab:class_aggregation}). For comparability with prior work~\cite{argoverse2,nuscenes,streammapnet,maptr,maptrv2}, we additionally evaluate the established three-class setting (pedestrian crossings, lane dividers, and road boundaries) for both RoIs.

We report class-wise average precision and \gls{map}, using distance thresholds of \(\{0.5, 1.0, 1.5\}\,\mathrm{m}\) and \(\{1.0, 1.5, 2.0\}\,\mathrm{m}\) for the smaller and larger RoIs, respectively. For the 53-class setting, we additionally report the KITScenes topology (TOP) score~\cite{openlanev2}.

\input{sec/tables_figures/53c_res}
\input{sec/tables_figures/qual_visu_new}

\paragraph{Implementation Details.}
\ifMemFusionCamera
    \ifMemFusionAID
        We evaluate StreamMapNet~\cite{streammapnet} and MemFusionMap~\cite{song2025memfusionmap} as camera-only baselines, StreamMapNet-CVS~\cite{cvs} with aerial imagery used only during training, and AerialFusionMapNet~\cite{afmn} and MemFusionMap + AID with camera--aerial fusion during training and inference.
    \else
        We evaluate StreamMapNet~\cite{streammapnet} and MemFusionMap~\cite{song2025memfusionmap} as camera-only baselines, StreamMapNet-CVS~\cite{cvs} with aerial imagery used only during training, and AerialFusionMapNet~\cite{afmn} with camera--aerial fusion during training and inference.
    \fi
\else
    \ifMemFusionAID
        We evaluate StreamMapNet~\cite{streammapnet} as a camera-only baseline, StreamMapNet-CVS~\cite{cvs} with aerial imagery used only during training, and AerialFusionMapNet~\cite{afmn} and MemFusionMap + AID~\cite{song2025memfusionmap} with camera--aerial fusion during training and inference.
    \else
        We evaluate StreamMapNet~\cite{streammapnet} as a camera-only baseline, StreamMapNet-CVS~\cite{cvs} with aerial imagery used only during training, and AerialFusionMapNet~\cite{afmn} with camera--aerial fusion during training and inference.
    \fi
\fi
We extend all methods from 2D to 3D map representations to leverage the rich 3D annotations of KITScenes, particularly for traffic lights and signs.
For AerialFusionMapNet, we reduce the aerial feature dimensionality to limit the storage requirements of the released DOP8 features, adapt the aerial branch's downsampling to match the required spatial resolution before fusion, and adopt simplified single-stage joint training.
Otherwise, we follow the original implementations, leaving further method-specific optimizations for future work.
Unless stated otherwise, we use DOP20 aerial imagery.

\paragraph{Results for Rich HD Map Construction.}
In \cref{tab:53class_results}, we report the performance of three state-of-the-art approaches.
While some semantic groups, like lane and road markings (LM/RM) benefit from distilling aerial information during training, groups that are hardly observable from above, like traffic lights and signs (TL/TS) even get detected less reliably.
In contrast, using aerial imagery during training and inference can yield a significant boost across nearly all classes, boosting overall mAP by 32\% and topology AP by 4.6 percentage points.

\paragraph{Effects of Increased Mapping Range.}
Comparing these numbers with \cref{tab:53class_extended_results}, the same evaluation at an extended \qtyproduct{100 x 50}{\meter} range, the relative benefit of using aerial imagery for both distillation and during inference grows significantly.
With aerial imagery available during inference, mAP more than doubles and topology AP improves by 11.7 percentage points, underlining the potential benefit of aerial imagery as an additional prior for long-range HD map construction.

\input{sec/tables_figures/53c_extended_res}
\paragraph{Effects of Higher Resolution.}
Additionally, to examine the influence of aerial imagery resolution, in \cref{tab:aerial_resolution_ablation}, we compare UNet++-based AerialFusionMapNet using imagery with ground sample distances of \cmppx{20} and \cmppx{8}.
The higher resolution can indeed improve rich HD map construction performance by up to 6.5 mAP (+36\%), demonstrating the potential of the published UNet++ features.
However, topology performance decreases at the extended range, suggesting that improved element detection does not necessarily translate into better topological consistency.
\input{sec/tables_figures/abl_dop20vs10}

\input{sec/tables_figures/3c_res}
\paragraph{Results for 3-Class HD Map Construction.}
Finally, to make results on KITScenes Aerial, featuring rich production-ready HD maps, comparable with those on previous datasets, in \cref{tab:3class_results}, we report results for all compared methods on the three established classes of map elements.
The findings from the rich 53-class setup are confirmed:
While distillation only yields significant benefits at long perception ranges, utilizing aerial imagery during training and inference improves HD map perception performance across the board.

\subsection{Aerial HD Map Construction}

With its rich HD map designed and ready for autonomous driving, KITScenes Aerial is a challenging benchmark for aerial-only vector map construction and rasterized HD map segmentation.

This is underlined by the results in \cref{tab:53class_results}, which report results from an AerialFusionMapNet variant using only aerial imagery.
It achieves 11.9 mAP on the 53-class KITScenes benchmark, approaching the camera-only baseline (13.6 mAP) and even outperforming it on Lane Markings (19.1 vs. 14.9 AP) and Lane Centerlines (32.2 vs. 30.2 AP).
The remaining gap to onboard and fused variants, particularly in topology, highlights the difficulty of direct fine-grained vector map construction from aerial imagery alone.

\begin{figure}[t]
    \centering
    \includegraphics[width=0.24\linewidth]{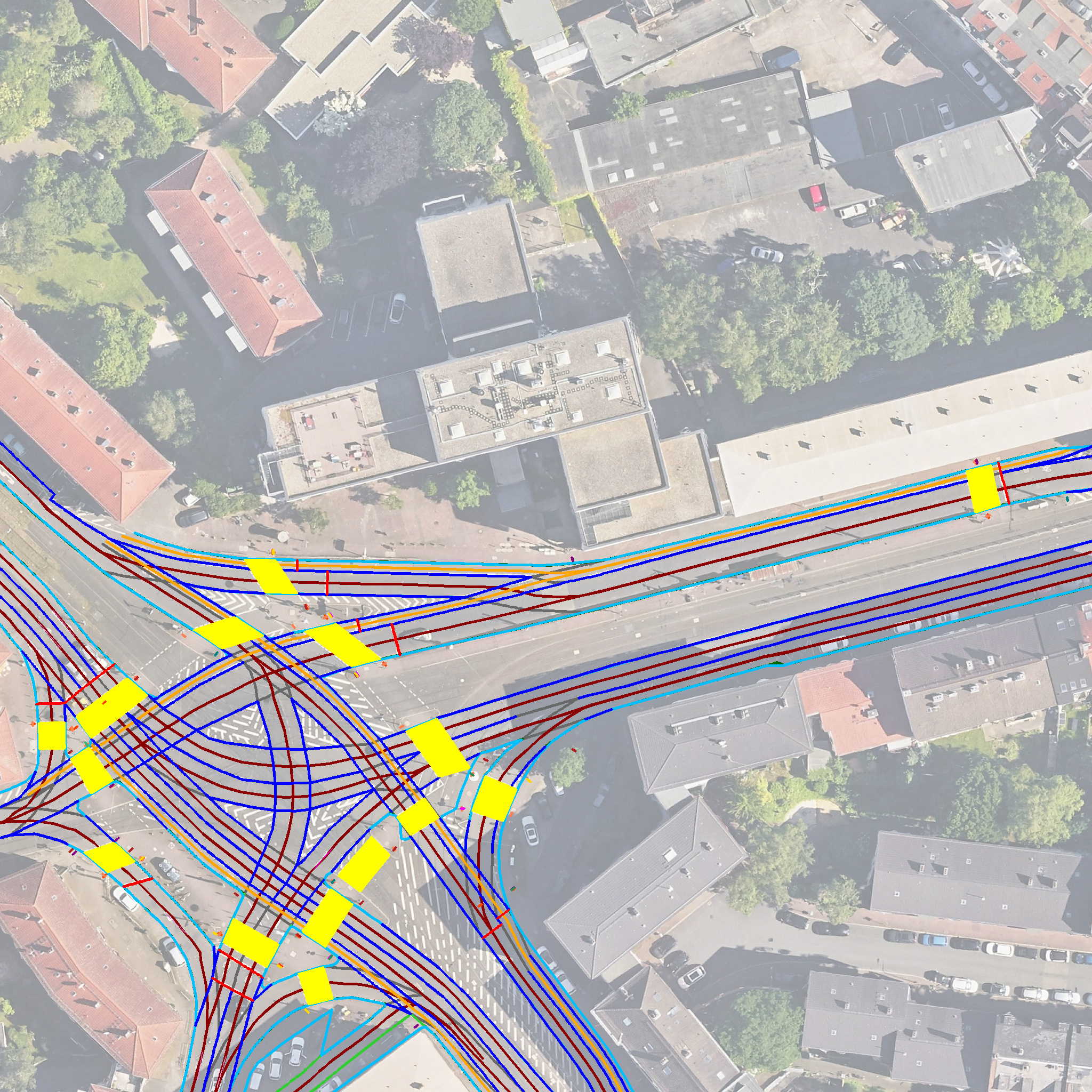}
    \hfill
    \includegraphics[width=0.24\linewidth]{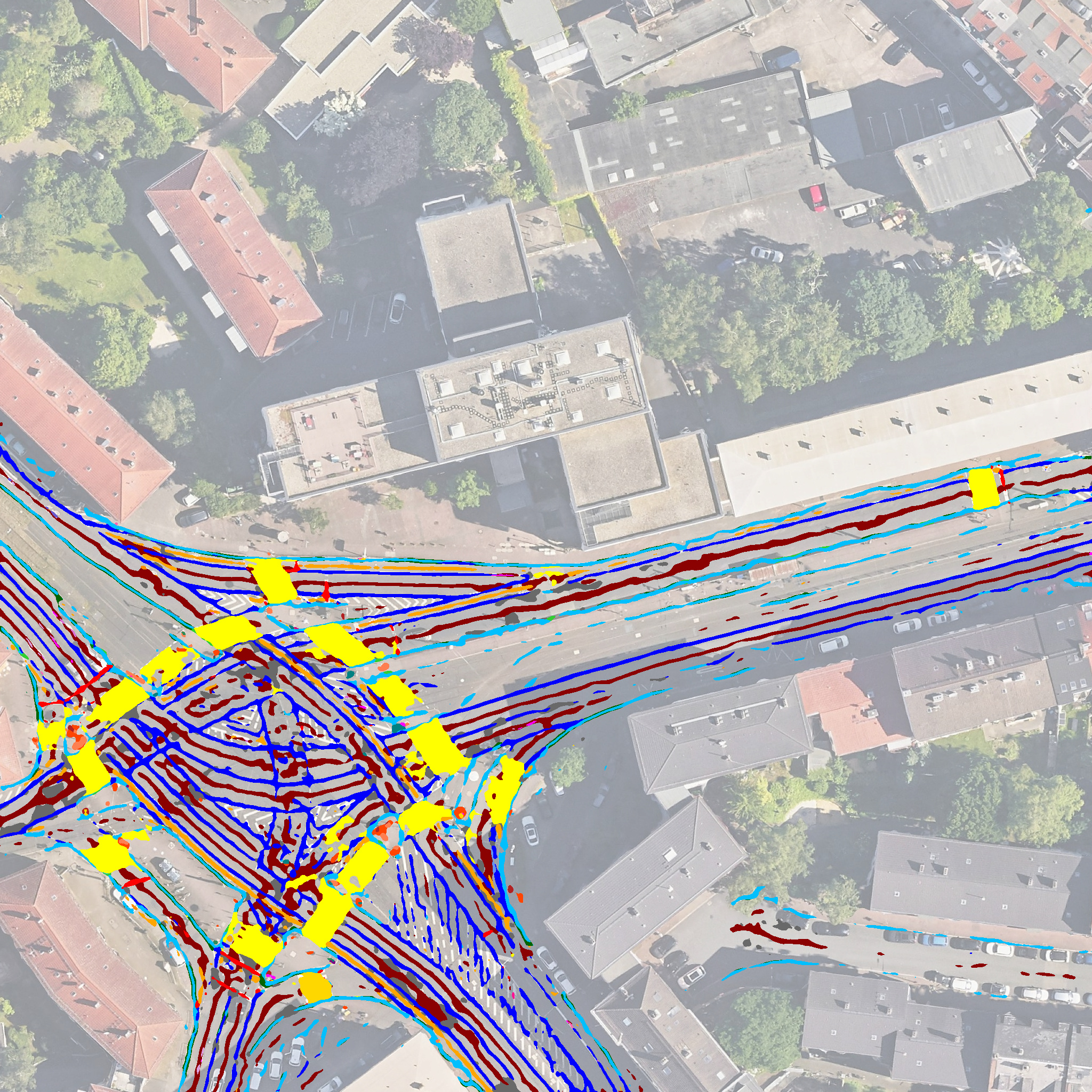}
    \hfill
    \includegraphics[width=0.24\linewidth]{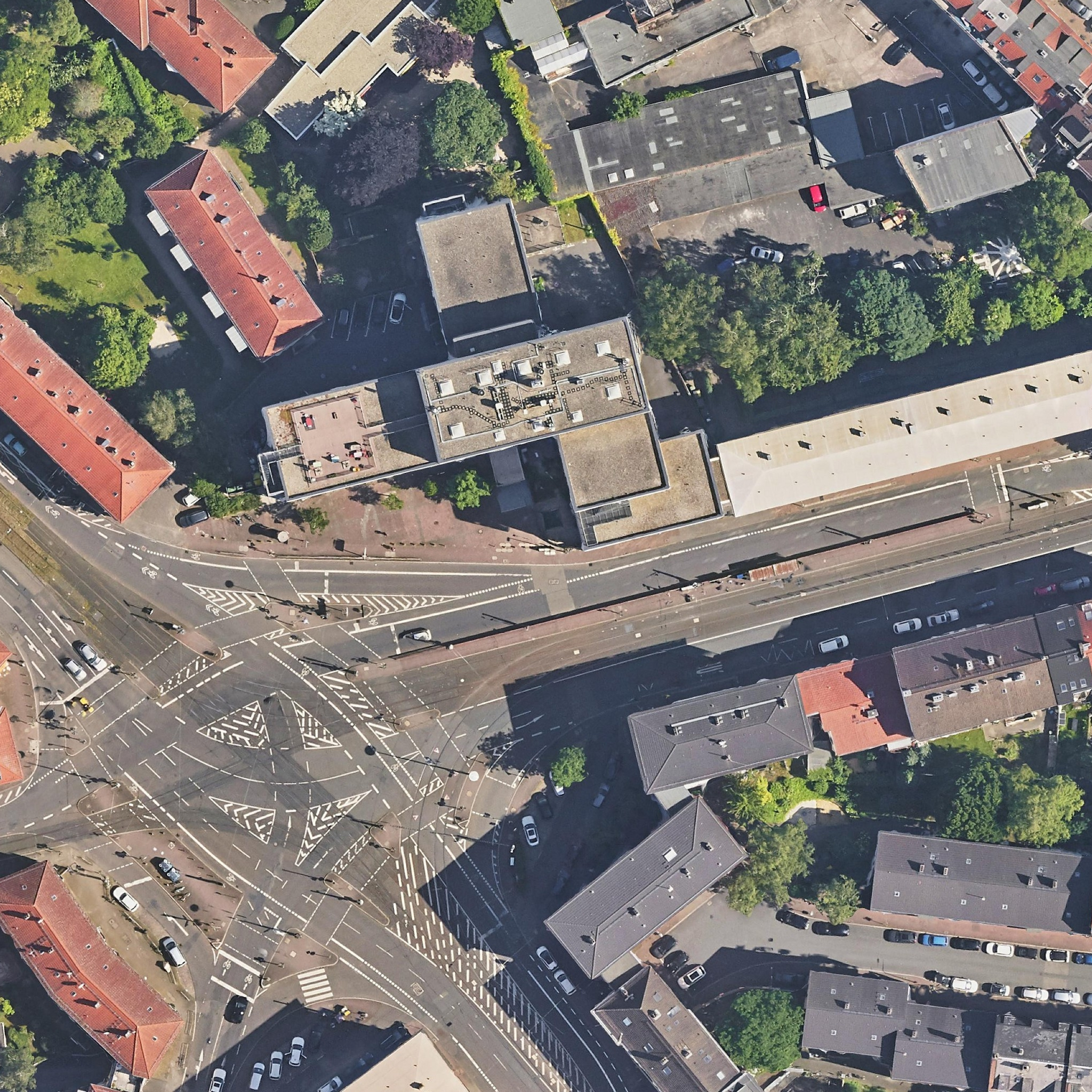}
    \hfill
    \includegraphics[width=0.24\linewidth]{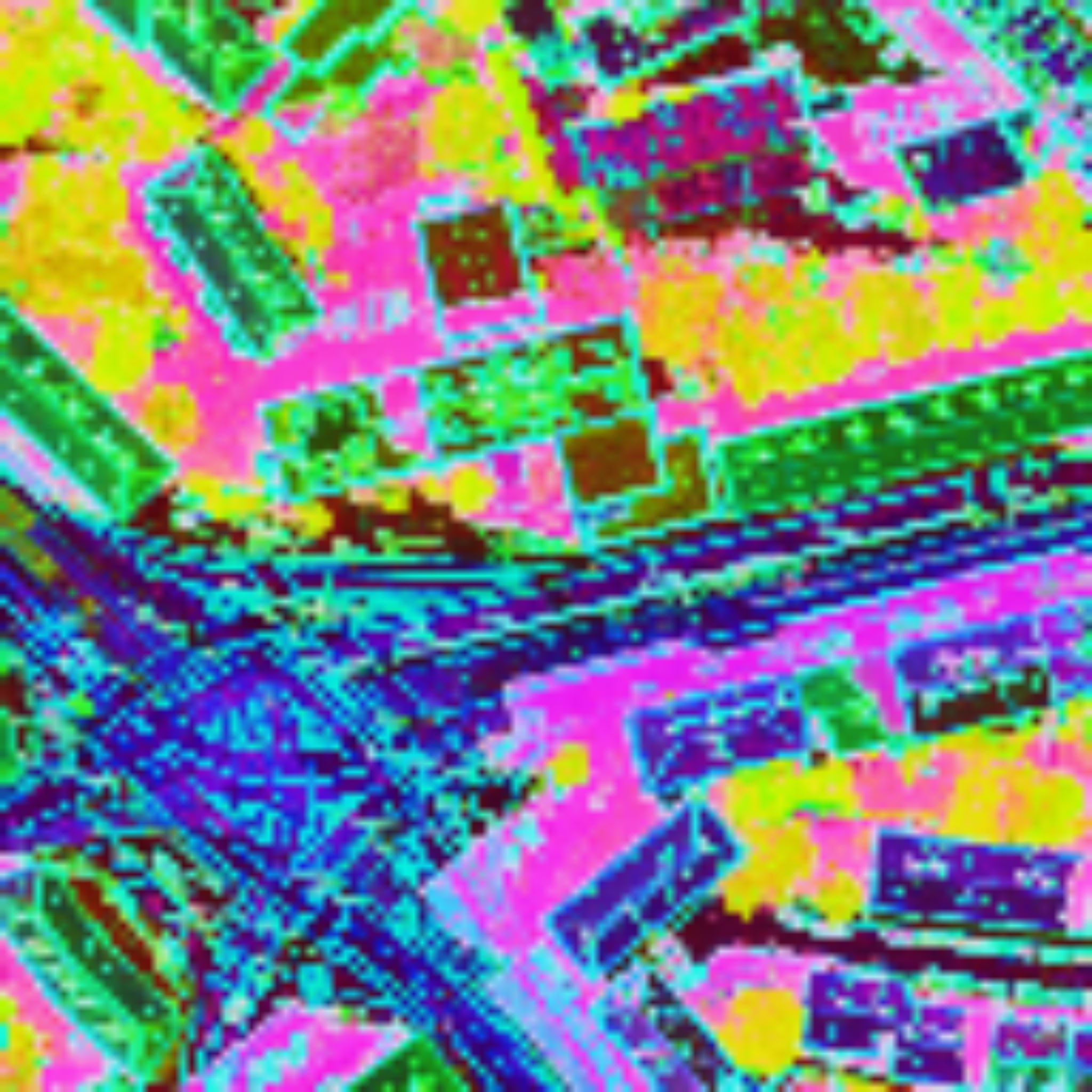}

    \includegraphics[width=0.24\linewidth]{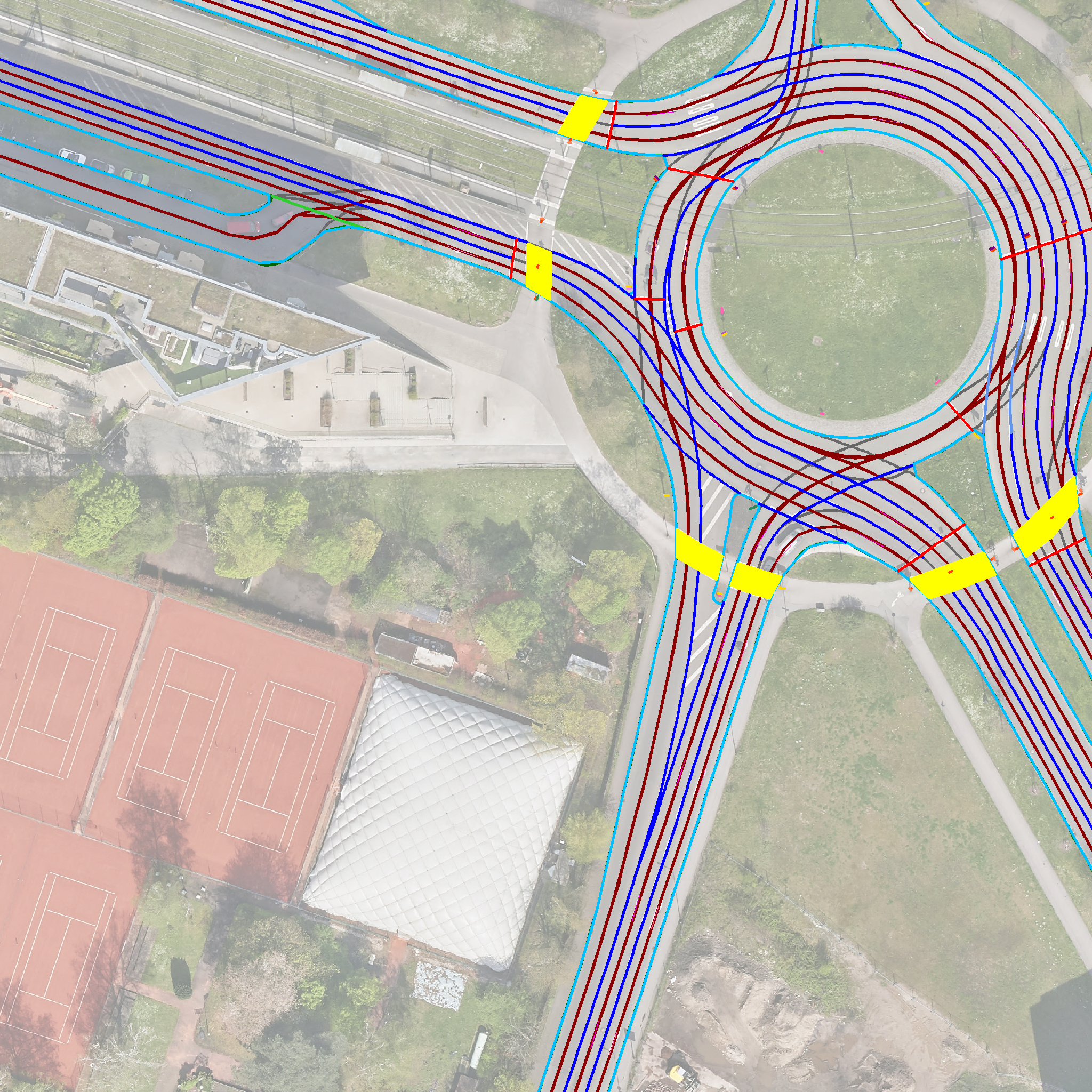}
    \hfill
    \includegraphics[width=0.24\linewidth]{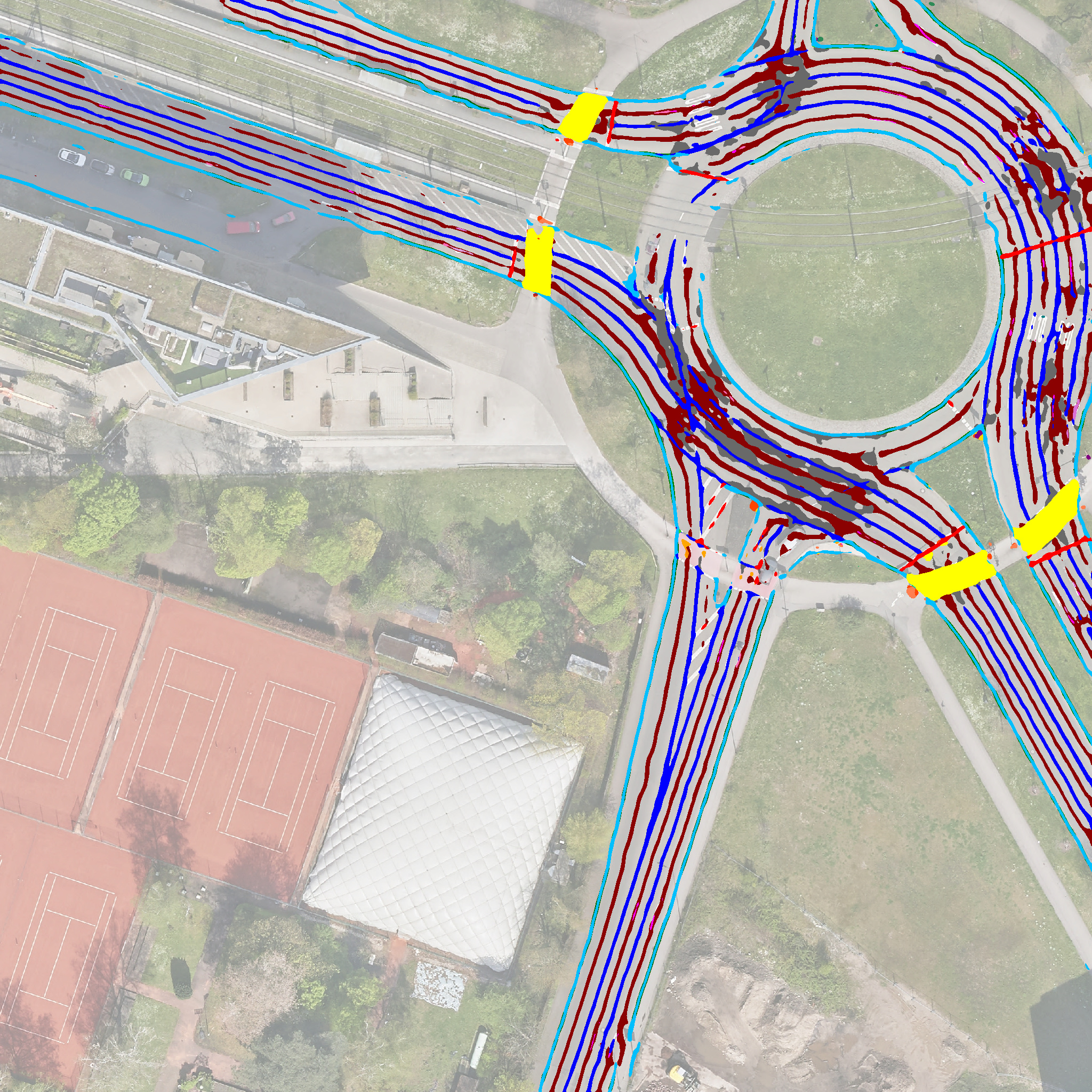}
    \hfill
    \includegraphics[width=0.24\linewidth]{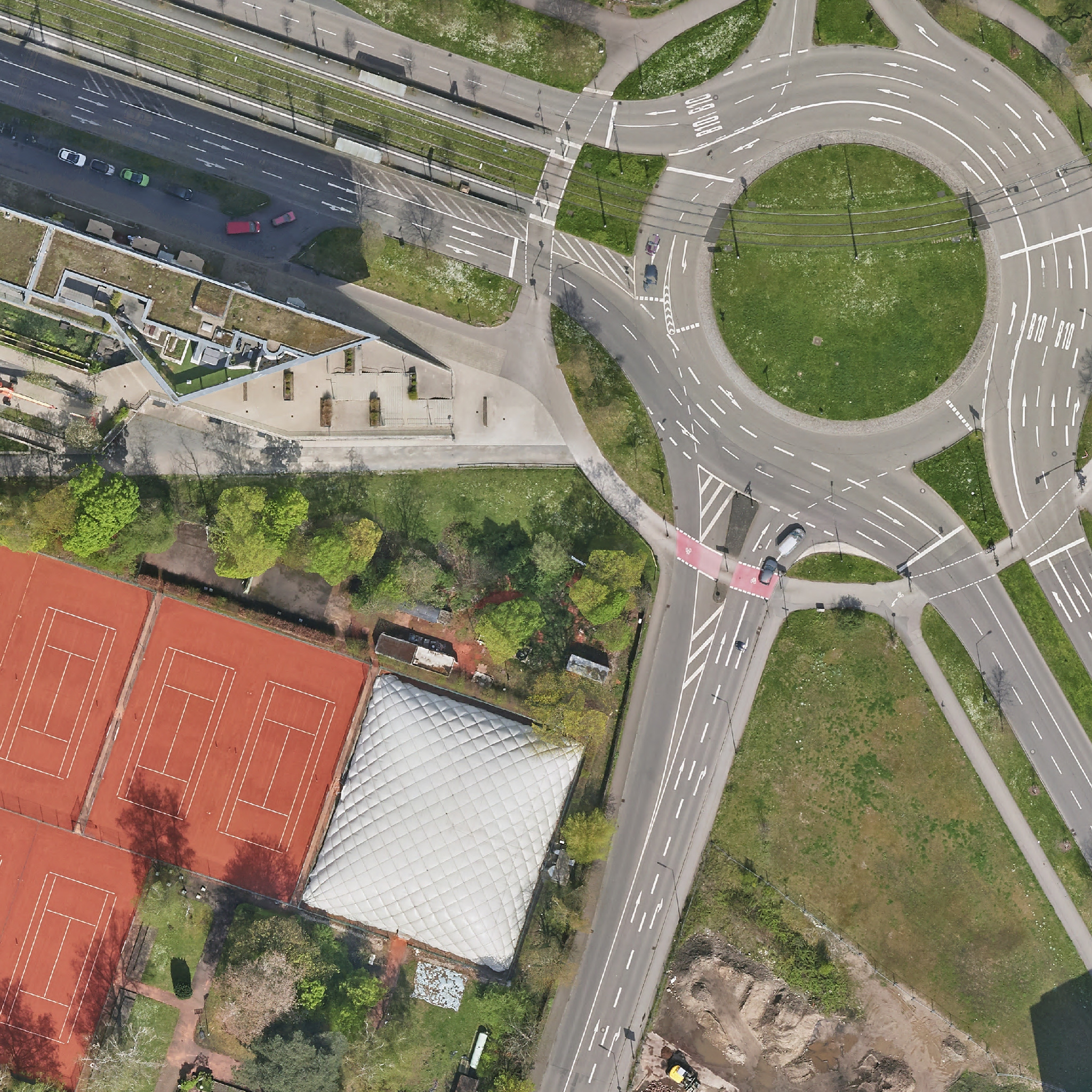}
    \hfill
    \includegraphics[width=0.24\linewidth]{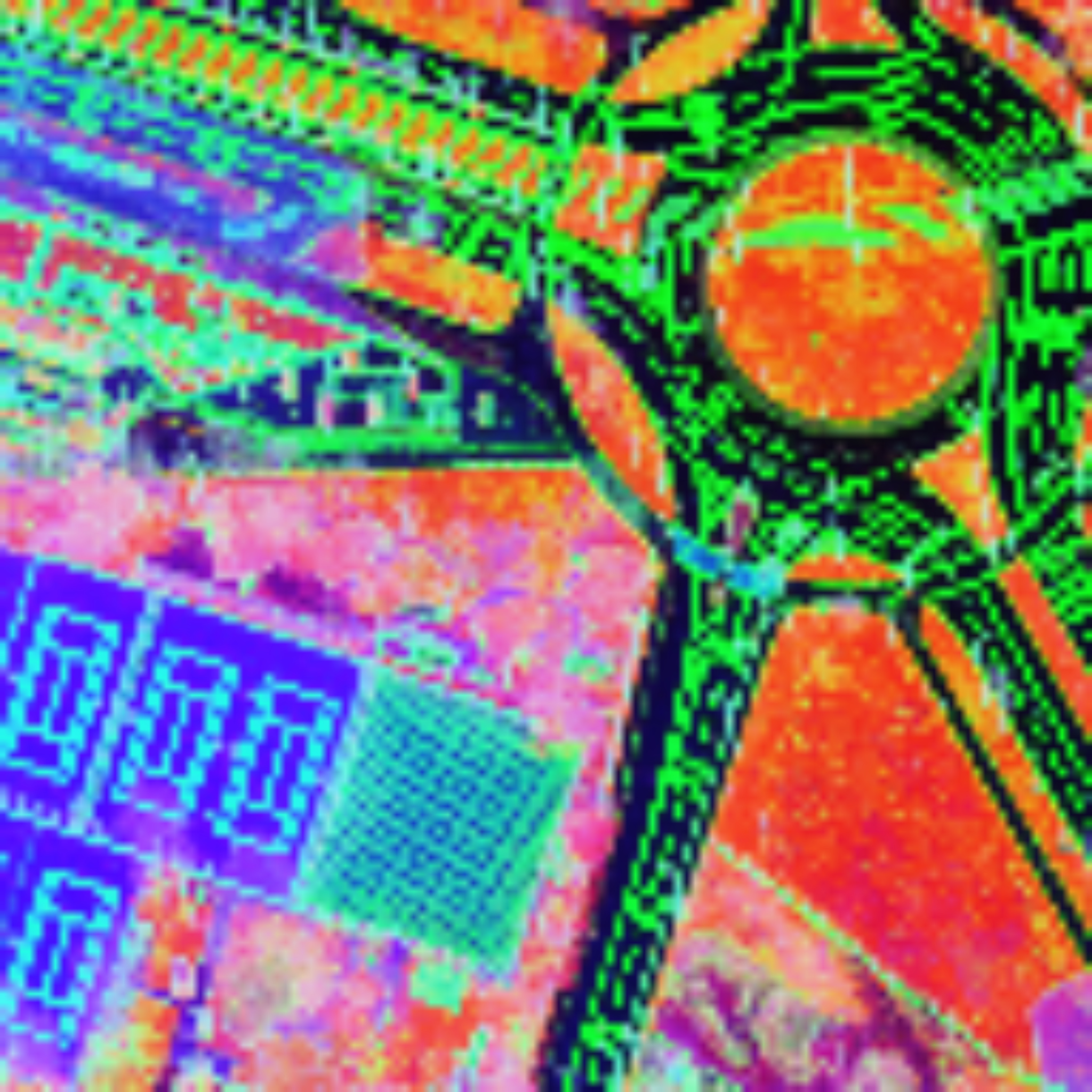}
    
    \caption{Rasterized HD map segmentation results by a DPT~\cite{dpt_vit_for_dense_prediction} decoder trained and inferred directly on the released DINOv3 features. Left to right: ground truth, prediction, higher-resolution aerial imagery (DINOv3 input), PCA of DINOv3 features.}
    \label{fig:aerial_segmentation_example_mosaic}
\end{figure}

\begin{figure}[t]
    \centering
    \includegraphics[width=0.32\linewidth]{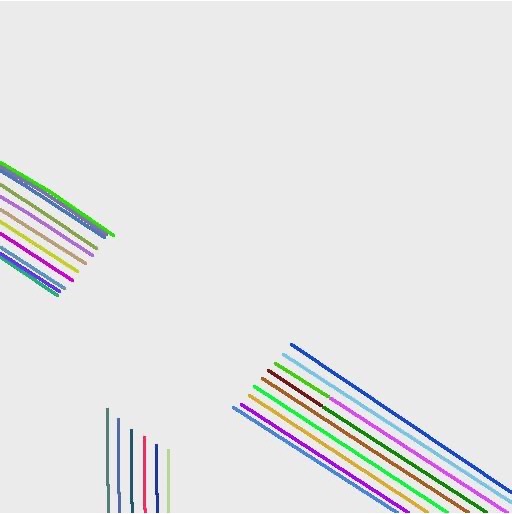}
    \hfill
    \includegraphics[width=0.32\linewidth]{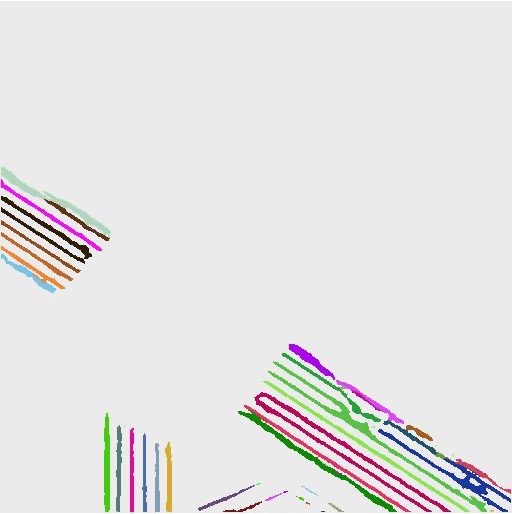}
    \hfill
    \includegraphics[width=0.32\linewidth]{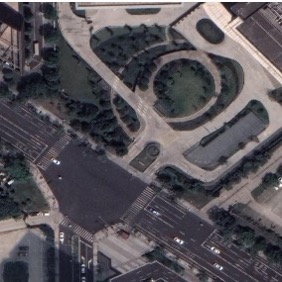}
    \caption{OpenSatMap~\cite{opensatmap} HD map segmentation representation and qualitative segmentation results. Although OpenSatMap closely relates to our aerial HD map construction task, the map representation, aerial image quality, and benchmark task drastically differ from ours. Left to right: ground truth, prediction, aerial image. Plots and image from~\cite{opensatmap}.}
    \label{fig:opensatmap_comparison}
\end{figure}

In \cref{fig:aerial_segmentation_example_mosaic}, we demonstrate the suitability of the released high-resolution DINOv3 encoded imagery features for fine-grained mapping related tasks.
For this, city-scale maps are rasterized onto \cmppx{8} tiles.
Tiles that intersect a KITScenes Multimodal validation scene map element are validation tiles, the rest of the tiles are training tiles, so the evaluation stays geographically split and aligned with the KITScenes Multimodal split.
For evaluation, the computed DINOv3 SAT ViT-L/16~\cite{simeoni2025dinov3} features are frozen as published.
Only a dense DPT~\cite{dpt_vit_for_dense_prediction} segmentation head is trained on the task of global rasterized HD map segmentation similar to HDMapNet~\cite{hdmapnet} using the 53 map element classes.
\cref{tab:opensatmap_compare} reports the segmentation results (mIoU) in the native fine-grained class label space and provides the OpenSatMap~\cite{opensatmap} segmentation results as a proxy comparative value, when the KITScenes Aerial labelspace is remapped to the OpenSatMap~\cite{opensatmap} label convention. \cref{fig:aerial_segmentation_example_mosaic} and \cref{fig:opensatmap_comparison} provide a visual representation of these metric results in the respective validation split data.

\begin{table*}[t]
\centering
\caption{Comparison of semantic HD map segmentation (mIoU in \%) on KITScenes Aerial versus OpenSatMap.
We compare a SegNeXt-tiny on the OpenSatMap val split (8 line classes: background, curb, virtual, and five
painted line types) against a frozen DINOv3 SAT ViT-L with a DPT head trained on 53 Lanelet2 classes.
On KITScenes, we report all native 53 classes as well as classes remapped\textsuperscript{*} to the 8-class OpenSatMap ontology.
}
\label{tab:opensatmap_compare}
\begin{tabular}{llcr}
\toprule
Method & Evaluation Data & Semantic Classes & mIoU \\
\midrule
SegNeXt-T~\cite{guo2022segnext} & OpenSatMap19~\cite{opensatmap} & 8 line classes & 28.71 \\
SegNeXt-T~\cite{guo2022segnext} & OpenSatMap20~\cite{opensatmap} & 8 line classes & 33.69 \\
DINOv3 SAT ViT-L~\cite{simeoni2025dinov3} + DPT~\cite{dpt_vit_for_dense_prediction} & KITScenes Aerial DINOv3 val & native 53 HD map classes & 8.58 \\
DINOv3 SAT ViT-L~\cite{simeoni2025dinov3} + DPT~\cite{dpt_vit_for_dense_prediction} & KITScenes Aerial DINOv3 val & 8 line classes (remapped\textsuperscript{*} to~\cite{opensatmap}) & 34.65 \\
\bottomrule
\end{tabular}
\vspace{0.7mm}
{\scriptsize
\begin{minipage}{0.96\textwidth}
\raggedright
\textsuperscript{*}\textit{Remapping:} Dashed/DashedSolid~$\to$~dashed, Solid/SolidSolid~$\to$~solid,
StopLine and remaining paint variants~$\to$~others, RoadBorder and Curbstones~$\to$~curb,
Virtual~$\to$~virtual, all other HD-map classes (centerlines, areas, signs,
arrows)~$\to$~background; unused types (ThickSolid, ShortDashed)
are excluded from mIoU.
\end{minipage}\par}
\end{table*}

In \cref{fig:opensatmap_comparison}, we provide qualitative examples of OpenSatMap~\cite{opensatmap}.
In comparison with \cref{fig:aerial_segmentation_example_mosaic}, they demonstrate the level of detail and completeness of the KITScenes Lanelet2 maps.
As shown in \cref{tab:opensatmap_compare}, this makes KITScenes Aerial a significantly more challenging task for holistic aerial HD map segmentation.
Specifically, we compare SegNeXt-tiny~\cite{guo2022segnext} on 8 lane classes against our frozen DINOv3 SAT ViT-L backbone with a DPT segmentation head trained on 53 classes.
Additionally, we remap a part of the classes to the OpenSatMap ontology.
This equalizes the \emph{metric}, but not the imagery
(German DOP vs.\ Google satellite), GSD (\cmppx{8} DINOv3 features vs.\ \cmppxrange{15}{30}), or label
method and format (Lanelet2 vs.\ custom OpenSatMap).
Despite using a vastly more capable model, segmentation performance is almost identical between both datasets.

\subsection{Cross-View Localization}
\label{sec:cross_view_setup}

The third downstream application enabled by KITScenes Aerial is fine-grained cross-view localization~\cite{zhu2021vigor,shi2020beyond,fg2}.
Following established cross-area evaluation protocols~\cite{zhu2021vigor,shi2020beyond}, \cref{tab:fg2_localization} reports FG$^2$~\cite{fg2} localization performance using the geographically disjoint KITScenes training and validation splits.
We additionally evaluate the original KITTI~\cite{kitti}-trained model zero-shot on KITScenes Aerial to assess cross-dataset transfer.

\begin{table}[t]
\centering
\caption{FG$^2$~\cite{fg2} cross-view localization results under cross-area evaluation, reporting translational and orientation error on the val split.
KITTI denotes the reported result from~\cite{fg2}. For KITScenes Aerial,
zero-shot uses the KITTI-trained model without adaptation, while fine-tuned denotes the model adapted on the KITScenes training split.}
\label{tab:fg2_localization}
\setlength{\tabcolsep}{3pt}
\begin{tabular}{lcccc}
\toprule
& \multicolumn{2}{c}{Trans. [m] $\downarrow$}
& \multicolumn{2}{c}{Orient.  [$^\circ$] $\downarrow$} \\
Setting & Mean & Median & Mean & Median \\
\midrule
KITTI             & 7.45 & 4.03 & 3.33 & 1.88 \\
\midrule
KITScenes (zero-shot)        & 11.38 & 9.00 & 4.27 & 2.89 \\
KITScenes (fine-tuned)       & 4.56 & 2.59 & 2.76 & 1.88 \\
\bottomrule
\end{tabular}
\end{table}

Despite the additional cross-dataset domain shift, the KITTI-trained model already transfers to KITScenes Aerial without adaptation. 
Fine-tuning on KITScenes Aerial substantially improves localization performance under the geographically separated cross-area evaluation.

%% file: sec/tables_figures/53c_subgroup_explained.tex
\begin{table}[t]
\centering
\caption{For reporting, we aggregate the 53 KITScenes map-element classes into six categories, listed with representative entries.}
\label{tab:class_aggregation}
\footnotesize
\setlength{\tabcolsep}{0pt}
\renewcommand{\arraystretch}{1.05}
\begin{tabularx}{\columnwidth}{
    >{\raggedright\arraybackslash}X
    >{\raggedright\arraybackslash}X
    >{\raggedright\arraybackslash}X}
\toprule
Lane Markings (LM) &
Lane Centerlines (LC) &
Road Infrastructure (RI) \\
\midrule
\texttt{road\_border} &
\texttt{centerline} &
\texttt{fence} \\

\texttt{dashed} &
{\footnotesize\texttt{bike\_centerline}} &
\texttt{building} \\

\ldots& & \ldots \\
\midrule
Traffic Lights (TL) &
Traffic Signs (TS) &
Road Markings (RM) \\
\midrule
\texttt{tl\_car} &
\texttt{ts\_stop} &
\texttt{bike\_symbol} \\

\texttt{tl\_bike} &
\texttt{ts\_yield} &
\texttt{bus\_symbol} \\

\ldots & \ldots & \ldots \\
\bottomrule
\end{tabularx}
\end{table}

%% file: sec/tables_figures/53c_res.tex
\begin{table*}[t]
\centering
\caption{
Online \gls{hd} map construction performance on the full KITScenes dataset
using the 53-class label space, evaluated within a
\qtyproduct{60 x 30}{\meter} region of interest.
For readability, class-wise AP is grouped into six semantic categories:
Lane Markings (LM), Lane Centerlines (LC), Road Infrastructure (RI),
Traffic Lights (TL), Traffic Signs (TS), and Road Markings (RM).
We additionally report the KITScenes topology score (Topo).
C: camera images, AID: aerial image data.
}
\label{tab:53class_results}
\setlength{\tabcolsep}{4pt}
\begin{tabular}{
l
|
c
c
|
S[table-format=2.1]
S[table-format=2.1]
S[table-format=2.1]
S[table-format=2.1]
S[table-format=2.1]
S[table-format=2.1]
|
S[table-format=2.1]
S[table-format=2.1]
}
\toprule
\multicolumn{1}{c|}{Method} &
\multicolumn{1}{c}{Training} &
\multicolumn{1}{c|}{Inference} &
{AP$_{LM}$} &
{AP$_{LC}$} &
{AP$_{RI}$} &
{AP$_{TL}$} &
{AP$_{TS}$} &
{AP$_{RM}$} &
{\acrshort{map}$_{53}$} &
{{$\mathrm{AP}_{\mathrm{Topo}}$}} \\
\midrule
StreamMapNet \cite{streammapnet}
    & C & C
    & 14.9 & 30.2 & 22.1 & 18.2 & 5.9 & 17.1 & 13.6 & 40.8 \\

StreamMapNet-CVS \cite{cvs}
    & C + AID
    & C
    & 17.1 & 31.6 & 22.2 & 15.2 & 2.6 & 20.0 & 13.3 & 40.6 \\

AerialFusionMapNet \cite{afmn}
    & C + AID
    & C + AID
    & 24.3 & 35.9 & 27.0 & 19.3 & 4.6 & 27.3 & 18.0 & 45.4 \\
\midrule
AerialFusionMapNet
    & AID
    & AID
    & 19.1 & 32.2 & 20.3 & 7.3 & 1.0 & 18.0 & 11.9 & 30.0 \\
\bottomrule
\end{tabular}
\end{table*}

%% file: sec/tables_figures/qual_visu_new.tex
\begin{figure*}[t]
    \centering

    \setlength{\tabcolsep}{2.5pt}
    \renewcommand{\arraystretch}{1.0}

    \begin{tabular}{
        @{}
        >{\centering\arraybackslash}m{0.242\textwidth}
        >{\centering\arraybackslash}m{0.242\textwidth}
        >{\centering\arraybackslash}m{0.242\textwidth}
        >{\centering\arraybackslash}m{0.242\textwidth}
        @{}
    }
        \small\textbf{Ground Truth} &
        \small\textbf{StreamMapNet} &
        \small\textbf{AerialFusionMapNet} &
        \small\textbf{\shortstack{AerialFusionMapNet\\Camera View}}
        \\[1.2mm]

        \includegraphics[width=\linewidth]{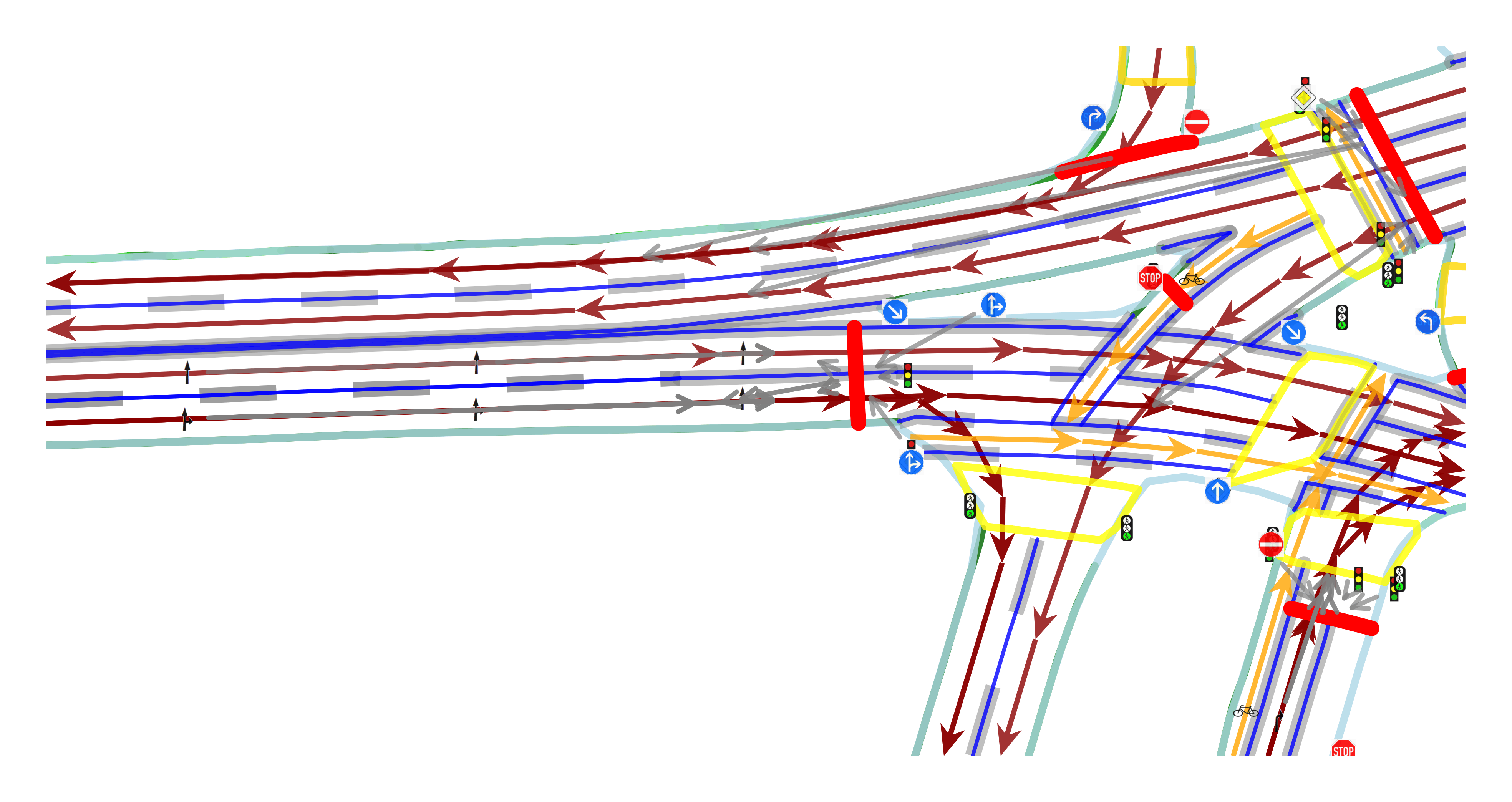} &
        \includegraphics[width=\linewidth]{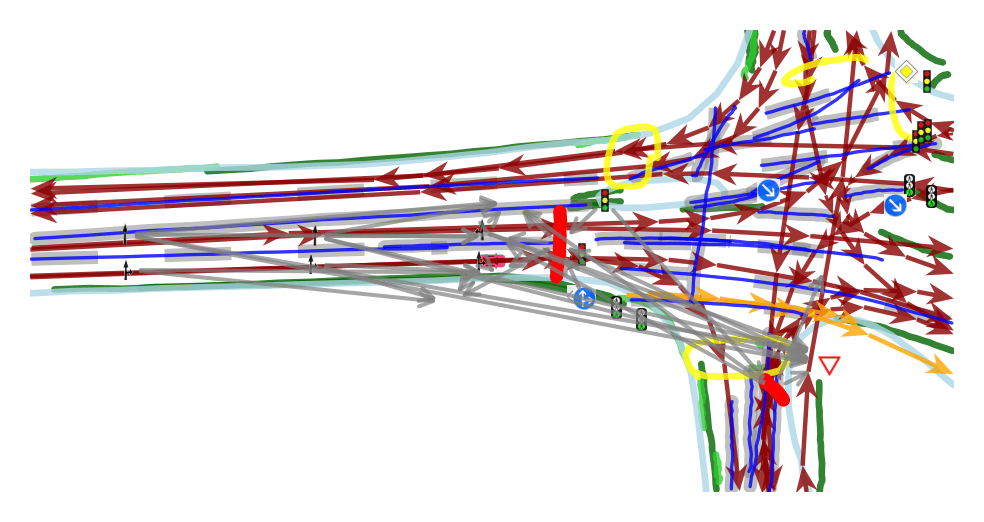} &
        \includegraphics[width=\linewidth]{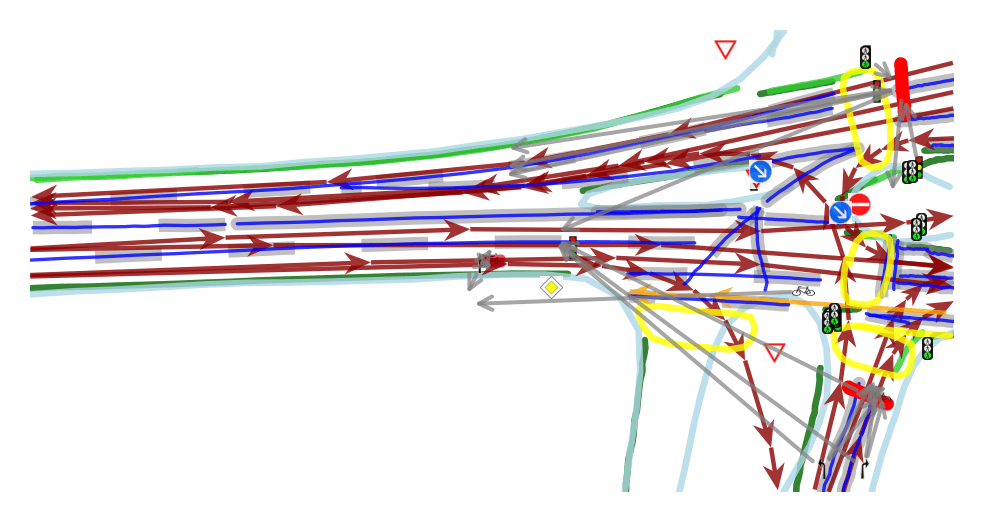} &
        \includegraphics[width=\linewidth]{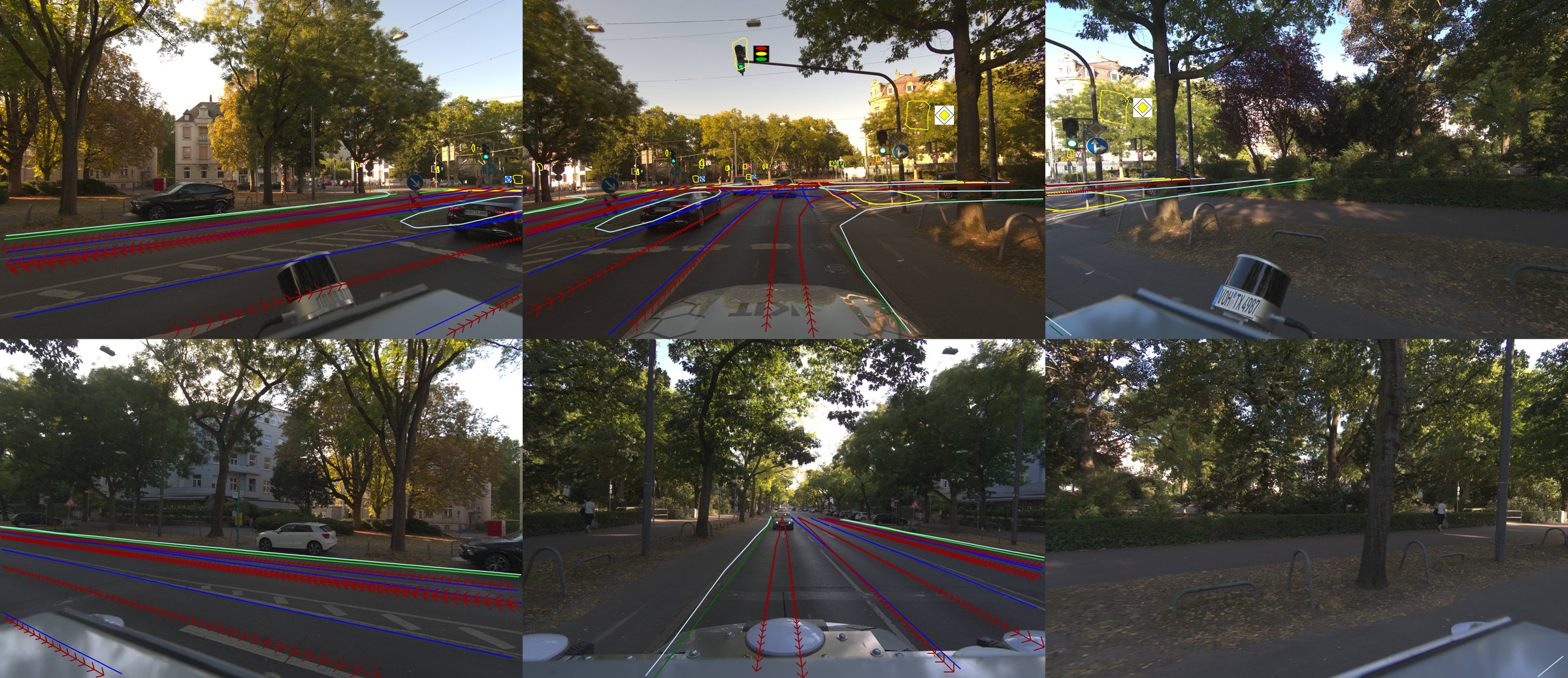}
        \\[1.2mm]

    \end{tabular}

    \vspace{-1mm}

    \caption{
        Qualitative comparison of online HD map construction on KITScenes Aerial.
        Predictions from the onboard-only StreamMapNet baseline are compared with the aerial-fused AerialFusionMapNet.
        Camera views show the corresponding ground-level context with AerialFusionMapNet detections.
    }
    \label{fig:qualitative_mapping}
\end{figure*}

%% file: sec/tables_figures/53c_extended_res.tex
\begin{table}[t]
\centering
\caption{
Extended-range online \gls{hd} map construction performance on the full
KITScenes dataset using the 53-class label space and a
\qtyproduct{100 x 50}{\meter} region of interest.
We report mean average precision and the KITScenes topology score.
C: camera images, AID: aerial image data.
\textsuperscript{*}AID only used during training.
}
\label{tab:53class_extended_results}
\setlength{\tabcolsep}{5pt}
\begin{tabular}{
l
c
|
S[table-format=2.1]
S[table-format=2.1]
}
\toprule
\multicolumn{1}{c}{Method} &
\multicolumn{1}{c}{Input} &
{\acrshort{map}$_{53}$} &
{{$\mathrm{AP}_{\mathrm{Topo}}$}} \\
\midrule
StreamMapNet \cite{streammapnet}
    & C
    & 6.2 & 24.8 \\

StreamMapNet-CVS \cite{cvs}
    & C + AID\textsuperscript{*}
    & 7.6 & 24.8 \\

AerialFusionMapNet \cite{afmn}
    & C + AID\phantom{\textsuperscript{*}}
    & 14.3 & 36.5 \\
\bottomrule
\end{tabular}
\end{table}

%% file: sec/tables_figures/abl_dop20vs10.tex
\begin{table}[t]
\centering
\caption{
Impact of aerial image resolution on AerialFusionMapNet in the 53-class
KITScenes setting across two regions of interest. We report \gls{map}$_{53}$
and the KITScenes topology metric $\mathrm{AP}_{\mathrm{Topo}}$.
}
\label{tab:aerial_resolution_ablation}
\setlength{\tabcolsep}{6pt}
\begin{tabular}{
c
|
S[table-format=2.1]
S[table-format=2.1]
|
S[table-format=2.1]
S[table-format=2.1]
}
\toprule
& \multicolumn{2}{c|}{\qtyproduct{60 x 30}{\meter}}
& \multicolumn{2}{c}{\qtyproduct{100 x 50}{\meter}} \\
\multicolumn{1}{c|}{Aerial GSD}
& {\gls{map}$_{53}$}
& {{$\mathrm{AP}_{\mathrm{Topo}}$}}
& {\gls{map}$_{53}$}
& {{$\mathrm{AP}_{\mathrm{Topo}}$}} \\
\midrule
\qty{20}{\cm\per\pixel} & 18.0 & 45.4 & 14.3 & 36.5 \\
\qty{8}{\cm\per\pixel}  & 24.5 & 44.4 & 17.8 & 22.4 \\
\bottomrule
\end{tabular}
\end{table}

%% file: sec/tables_figures/3c_res.tex
\begin{table*}[t]
\centering
\caption{
Online \gls{hd} map construction results for the previously established three-class protocol.
{C}: camera images, {AID}: aerial image data.
}
\label{tab:3class_results}

\setlength{\tabcolsep}{6pt}

\begin{tabular}{
l
|
c
c
|
S[table-format=2.1]
S[table-format=2.1]
S[table-format=2.1]
S[table-format=2.1]
|
S[table-format=2.1]
S[table-format=2.1]
S[table-format=2.1]
S[table-format=2.1]
}
\toprule

\multicolumn{1}{c}{} &
\multicolumn{1}{c}{} &
\multicolumn{1}{c}{} &
\multicolumn{4}{c}{\qtyproduct{60 x 30}{\meter}} &
\multicolumn{4}{c}{\qtyproduct{100 x 50}{\meter}} \\

\cmidrule(lr){4-7}
\cmidrule(lr){8-11}

\multicolumn{1}{c|}{Method} &
\multicolumn{1}{c}{Training} &
\multicolumn{1}{c|}{Inference} &
{AP$_{\text{ped}}$} &
{AP$_{\text{div}}$} &
{AP$_{\text{bnd}}$} &
{\gls{map}} &
{AP$_{\text{ped}}$} &
{AP$_{\text{div}}$} &
{AP$_{\text{bnd}}$} &
{\gls{map}} \\

\midrule

StreamMapNet \cite{streammapnet}
& C & C
& 21.1 & 34.1 & 37.6 & 30.9
& 16.8 & 18.8 & 16.9 & 17.5 \\

MemFusionMap \cite{song2025memfusionmap}
& C & C
& 22.7 & 35.3 & 40.2 & 32.6 %
& 18.7 & 19.7 & 19.3 & 19.2 \\ %

StreamMapNet-\acrshort{cvs} \cite{cvs}
& C + AID & C
& 22.7 & 35.1 & 36.9 & 31.6
& 20.7 & 22.2 & 21.6 & 21.5 \\

AerialFusionMapNet \cite{afmn}
& C + AID & C + AID
& 37.9 & 44.6 & 41.4 & 41.3
& 51.1 & 37.4 & 32.8 & 40.4 \\

MemFusionMap \cite{song2025memfusionmap} + AID 
& C + AID & C + AID
& 45.4 & 46.2 & 42.7 & 44.7 %
& 58.9 & 42.9 & 36.4 & 46.0 \\ %

\bottomrule
\end{tabular}
\end{table*}

%% file: sec/06_disc.tex
\section{Limitations}

\paragraph{Temporal Consistency.}
The effect of temporal differences between ground-level and aerial observations cannot be isolated. To minimize such discrepancies, KITScenes Aerial provides a fixed snapshot of the raw aerial tiles selected to be as temporally close as possible to the KITScenes recordings. Nevertheless, changes in road infrastructure between the respective acquisition times may remain.

\paragraph{Higher-Resolution Imagery.}
The higher-resolution orthophotos used in this work cannot be redistributed as raw RGB imagery due to licensing restrictions. We therefore release precomputed neural representations that make this information available for learning-based experiments, while limiting methods that require direct access to the underlying imagery or alternative feature extractors.

\paragraph{Benchmark Extensibility.}
KITScenes Aerial provides representative baselines and standardized interfaces for the three research directions considered in this work, facilitating the integration of KITScenes into existing aerial and cross-view methods. The presented experiments establish initial reference points for these settings and provide a foundation for broader benchmarking in future work.

%% file: sec/07_concl.tex
\section{Conclusion}
We introduced KITScenes Aerial as an aerial extension of KITScenes Multimodal that combines spatially associated aerial imagery with synchronized onboard sensing, vehicle poses, and detailed Lanelet2 maps.
Together, KITScenes Aerial is the first dataset that covers both high-resolution aerial imagery and driving-ready HD maps.
This provides a common data basis for aerial-only map construction, aerial-assisted online HD mapping, and cross-view localization. 

Our experiments show that aerial context becomes particularly valuable for fine-grained and extended-range map construction, while higher-resolution aerial imagery provides further gains for detailed map elements.
Cross-view localization experiments additionally demonstrate the applicability of the dataset under geographically disjoint conditions.
Beyond the evaluated settings, the joint availability of aerial and ground-level observations opens further directions for annotation-efficient learning, such as using aerial predictions as pseudo-labels for semi-supervised map learning in otherwise unlabeled driving regions or exploiting cross-modal agreement to improve pseudo-label quality.

These findings establish KITScenes Aerial as a foundation for studying previously largely separate ground--aerial research directions and for developing new forms of cross-modal supervision.

\section*{Acknowledgment}
The authors gratefully acknowledge the scientific support and HPC resources provided by the Erlangen National High Performance Computing Center (NHR@FAU) of the Friedrich-Alexander-Universität Erlangen-Nürnberg (FAU) under the BayernKI project v153eb. BayernKI funding is provided by Bavarian state authorities.

The research leading to these results is partially funded by the German Federal Ministry for Economic Affairs and Energy within the project NXT GEN AI METHODS (Grant Agreement No. 19A23014), and by the Shift2SDV project (Grant Agreement No. 101194245), co-funded by the European Union through the Chips Joint Undertaking and its members. The authors gratefully acknowledge the computing time provided on the high-performance computer HoreKa by the National High-Performance Computing Center at KIT (NHR@KIT). This center is jointly supported by the Federal Ministry of Education and Research and the Ministry of Science, Research and the Arts of Baden-Württemberg, as part of the National High-Performance Computing (NHR) joint funding program. HoreKa is partly funded by the German Research Foundation (DFG). The authors also gratefully acknowledge the support of the Large Scale Data Facility (LSDF) at the Karlsruhe Institute of Technology, funded by the Ministry of Science, Research and the Arts of Baden-Württemberg and by the Federal Ministry of Education and Research.

%% file: sec/08_supp.tex
\clearpage
\onecolumn
\raggedbottom
\appendix

\renewcommand{\thesection}{\Alph{section}}
\renewcommand{\thesubsection}{\thesection.\arabic{subsection}}
\renewcommand{\thefigure}{\thesection.\arabic{figure}}
\renewcommand{\thetable}{\thesection.\arabic{table}}

\setcounter{figure}{0}
\setcounter{table}{0}
\suppressfloats[t]

\begin{center}
    {\LARGE\bfseries KITScenes Aerial}\\[0.25em]
    {\large Appendix}
\end{center}

\vspace{1em}

\section{Additional KITScenes Aerial Examples}

\cref{fig:additional_samples_1}
shows additional
KITScenes Aerial samples across diverse driving scenes.

\input{sec/tables_figures/more_teaser_samples}

\FloatBarrier

\section{Additional Dataset Details}

\cref{fig:public_vs_municipal_imagery} compares
the publicly available DOP20 imagery with
higher-resolution licensed orthophotos.
\cref{tab:map_elem_to_category} lists the
53 map elements used for map construction and their
aggregation into six semantic categories for reporting.

\input{sec/tables_figures/sup_map_elemtents}
\input{sec/tables_figures/public_vs_municipal_fig}
\FloatBarrier

\section{Additional Experimental Results}
To complement the aggregate results reported in the main paper,
\cref{tab:subgroup_ap_gsd_comparison} provides a breakdown of AerialFusionMapNet performance across the six KITScenes
reporting categories for both aerial resolutions and regions of interest.
Higher-resolution imagery improves overall mAP in both settings,
with the largest gains for Road Markings
($+20.7$ and $+17.6$\,pp, respectively).
However, topology performance slightly decreases for the smaller
RoI and drops substantially for the extended range,
suggesting that improved element detection does not necessarily
translate into more accurate topological relationships.

\input{sec/tables_figures/appendix_20vs8inklsubgr}

\FloatBarrier

\clearpage
\subsection{Additional Qualitative Results}

\cref{fig:appendix_qualitative_mapping} presents
additional qualitative comparisons between StreamMapNet
and AerialFusionMapNet across diverse driving scenes,
complementing the examples in the main paper.

\input{sec/tables_figures/appendix_qual_visus}
\FloatBarrier

\section{Large-Scale Map Visualizations}

\cref{fig:stitched_large_scale_pred,fig:stitched_large_scale_pred2} show two
large-scale examples of rasterized HD map segmentation
predictions, each paired with its corresponding
ground truth visualization, \cref{fig:stitched_large_scale_gt,fig:stitched_large_scale_gt2}.

\input{sec/tables_figures/stitched_figures}

%% file: sec/tables_figures/more_teaser_samples.tex
\begin{figure}[H]
    \centering

    \includegraphics[width=\textwidth]
    {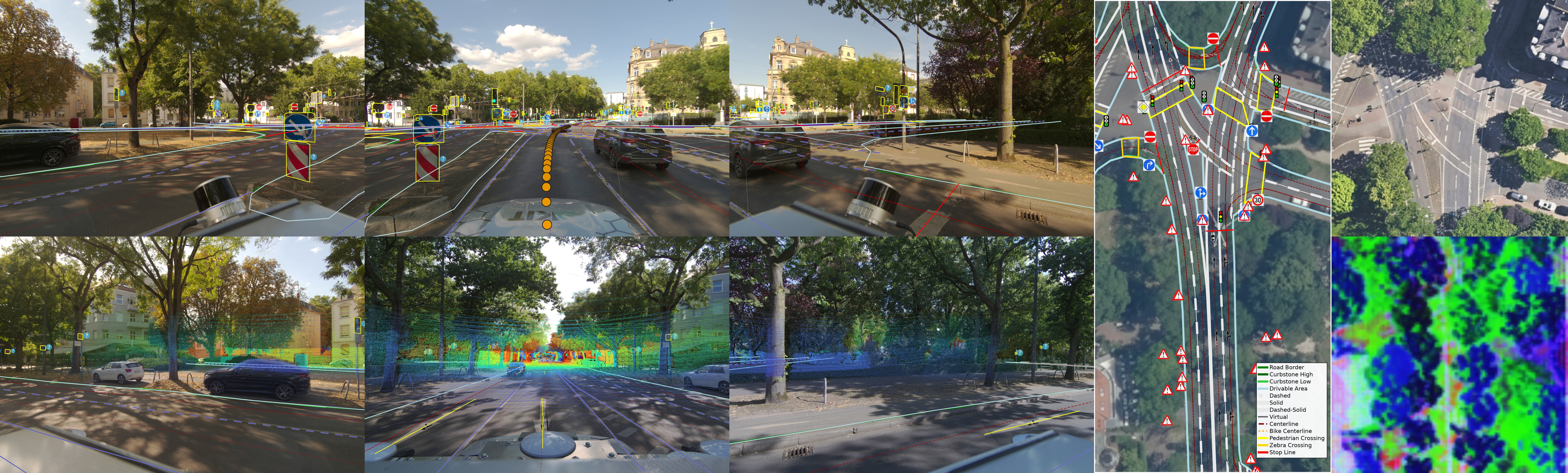}

    \vspace{1.5mm}

    \includegraphics[width=\textwidth]
    {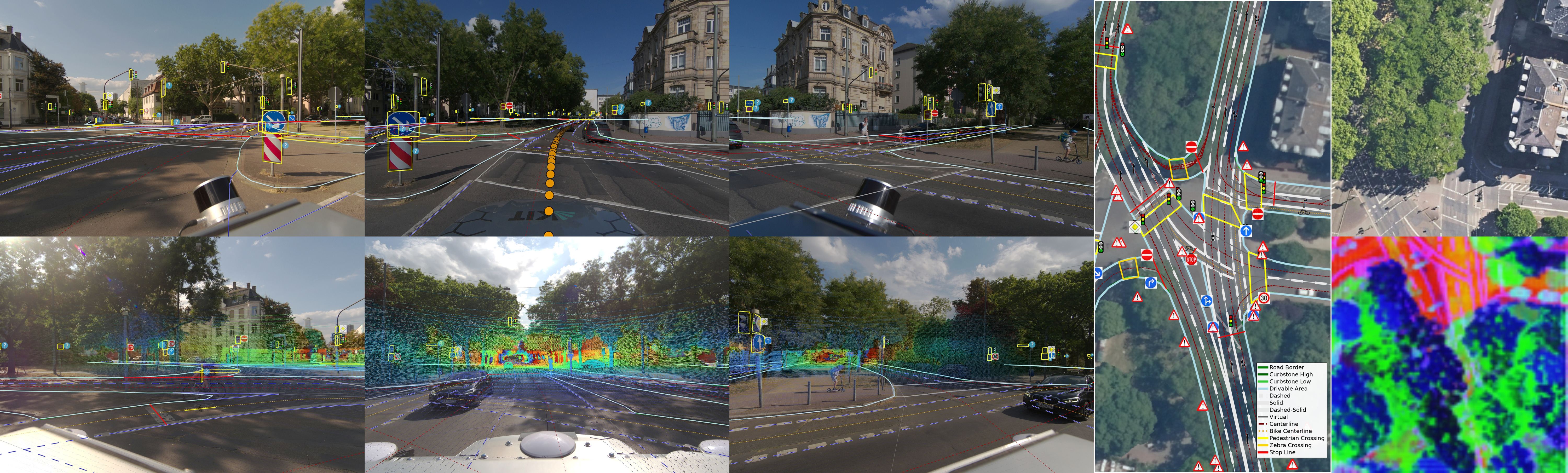}

    \vspace{1.5mm}

    \includegraphics[width=\textwidth]
    {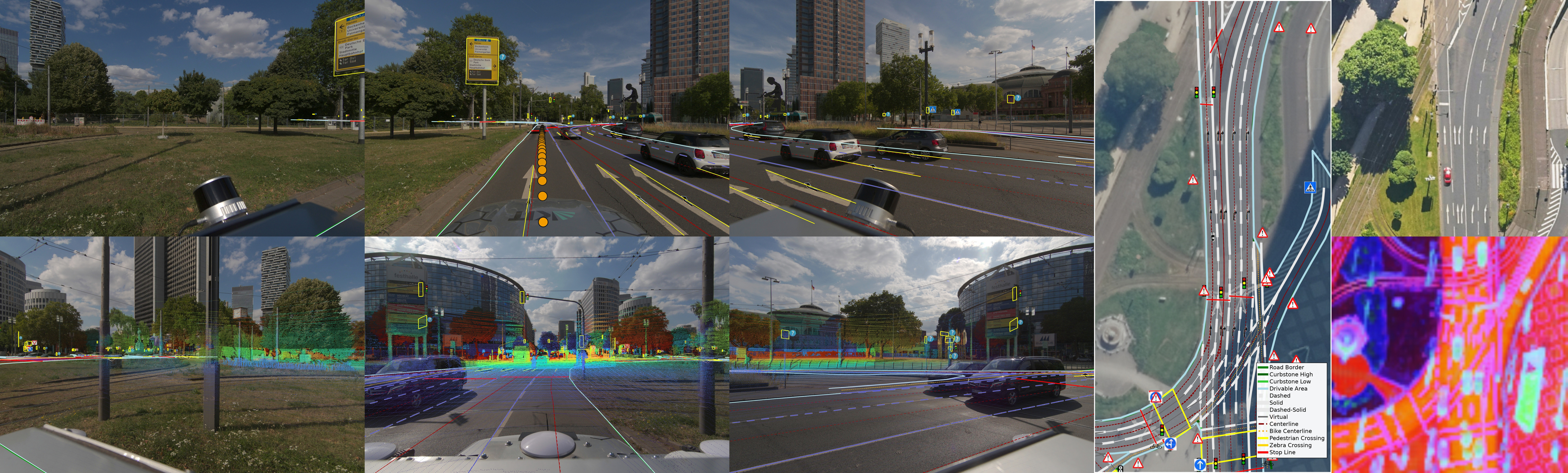}

    \caption{
        Additional KITScenes Aerial examples showing synchronized multimodal
        observations, HD map annotations, and corresponding aerial representations.
    }
    \label{fig:additional_samples_1}
\end{figure}

%% file: sec/tables_figures/sup_map_elemtents.tex
\begin{table*}[h]
  \centering
  \caption{Taxonomy of map elements used in the Lanelet2 map, grouped into six categories: Lane Markings (LM), Lane Centerlines (LC), Road Infrastructure (RI), Traffic Lights (TL), Traffic Signs (TS), and Road Markings (RM).}
  \renewcommand{\arraystretch}{1.2}
  \resizebox{0.9\textwidth}{!}{%
  \begin{tabular}{p{5cm} p{5cm} p{5cm}}
    \toprule
    \textbf{Lane Markings (LM)} & \textbf{Lane Centerlines (LC)} & \textbf{Road Infrastructure (RI)} \\
    \midrule
    \parbox[t]{4cm}{
      Road border \\ Dashed \\ Solid \\ Solid Solid \\
      Solid dashed \\ Dashed solid \\ Bike dashed \\
      Bike solid \\ Pedestrian crossing \\
      Zebra crossing \\ Stop line \strut
    }
    &
    \parbox[t]{4cm}{
      Centerline \\
      Bike centerline
    }
    &
    \parbox[t]{4cm}{
      Curbstone high \\ Curbstone low \\ Fence \\
      Building \\ Wall \\ Guardrail \\
      Drivable area \\ Divider
    }
    \\
    \bottomrule
  \end{tabular}%
  }

  \vspace{1em}

  \resizebox{0.9\textwidth}{!}{%
  \begin{tabular}{p{5cm} p{5cm} p{5cm}}
    \toprule
    \textbf{Traffic Lights (TL)} & \textbf{Traffic Signs (TS)} & \textbf{Road Markings (RM)} \\
    \midrule
    \parbox[t]{4cm}{
      Car \\ Bike \\
      Pedestrian \\ Misc
    }
    &
    \parbox[t]{4cm}{
      Stop \\ Yield \\ No entry \\
      Right of way \\ Priority road \\
      One way street \\ Roundabout \\
      Speed limit \\ Pedestrian crossing \\
      Turn right \\ Turn left \\
      Go straight \\ Go straight or right \\
      Go straight or left \\ Turn left or right \\
      Pass right \\ Pass left \strut
    }
    &
    \parbox[t]{4cm}{
      Arrow go straight \\ Arrow turn left \\
      Arrow turn right \\ Arrow go straight or left \\
      Arrow go straight or right \\
      Arrow turn left or right \\
      Bike symbol \\ Bus symbol \\
      Symbol 30 \\ Symbol 50 \\ Symbol 70 \strut
    }
    \\
    \bottomrule
  \end{tabular}%
  }
  \label{tab:map_elem_to_category}
\end{table*}

%% file: sec/tables_figures/public_vs_municipal_fig.tex
\begin{figure*}[ht]
    \centering
    \includegraphics[width=0.49\linewidth]{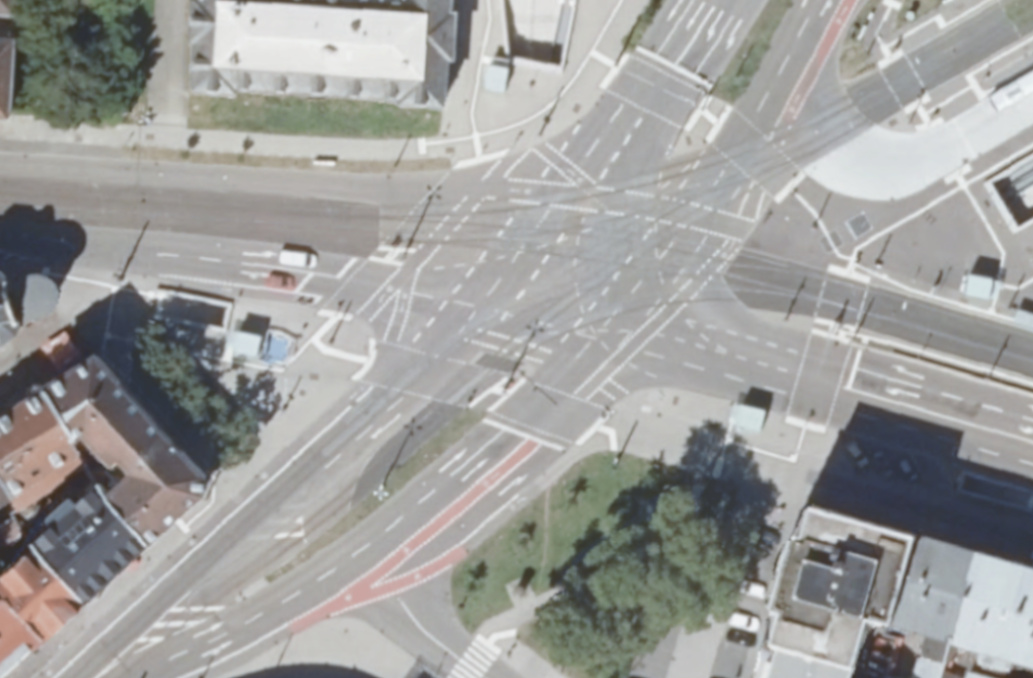}
    \hfill
    \includegraphics[width=0.49\linewidth]{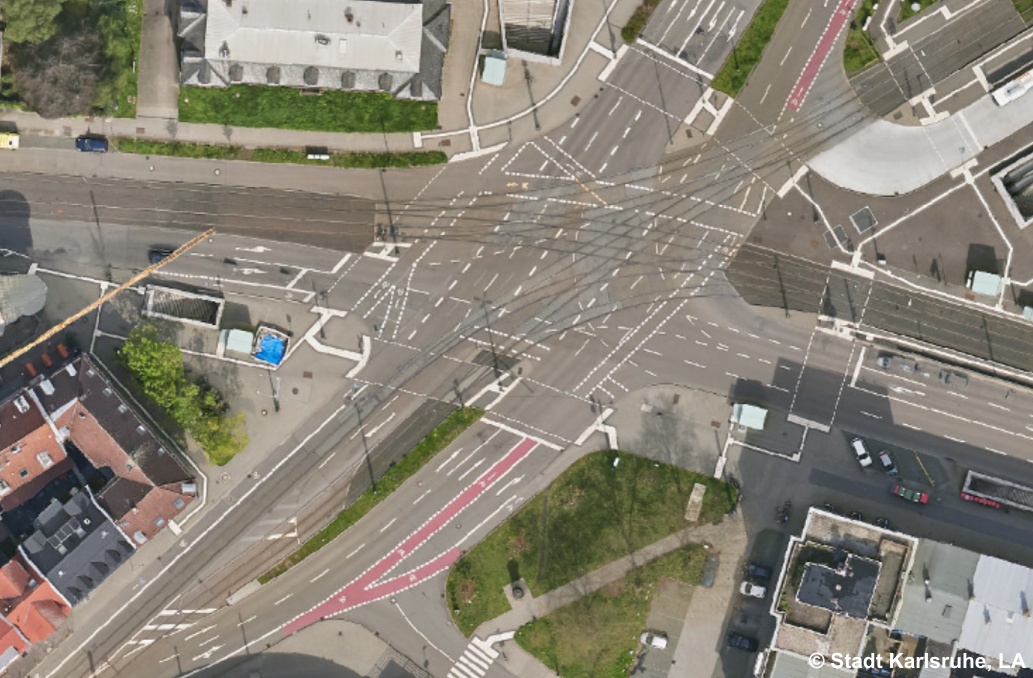}

    \caption{Qualitative comparison of country-level open aerial imagery (left) and high-quality municipal aerial imagery (right). We provide encoded neural features for higher-resolution aerial images of municipal quality.  Both images shown here are at a resolution of \qty{20}{\cm\per\pixel} due to privacy regulations.}
    \label{fig:public_vs_municipal_imagery}
\end{figure*}

%% file: sec/tables_figures/appendix_20vs8inklsubgr.tex
\begin{table*}[!htbp]
\centering
\caption{
Detailed AerialFusionMapNet results comparing 20\,cm and 8\,cm
GSD aerial imagery across the six KITScenes reporting categories
and both regions of interest.
Class grouping follows KITScenes (\cref{tab:map_elem_to_category}).
}
\label{tab:subgroup_ap_gsd_comparison}
\setlength{\tabcolsep}{4pt}
\begin{tabular}{
l
c
|
S[table-format=2.1]
S[table-format=2.1]
S[table-format=2.1]
S[table-format=2.1]
S[table-format=2.1]
S[table-format=2.1]
|
S[table-format=2.1]
S[table-format=2.1]
}
\toprule
{RoI (m)} &
{GSD} &
{AP$_{LM}$} &
{AP$_{LC}$} &
{AP$_{RI}$} &
{AP$_{TL}$} &
{AP$_{TS}$} &
{AP$_{RM}$} &
{\acrshort{map}} &
{{$\mathrm{AP}_{\mathrm{Topo}}$}} \\
\midrule
\multirow{2}{*}{$60 \times 30$}
    & 20\,cm
    & 24.3 & 35.9 & 27.0 & 19.3 & 4.6 & 27.3 & 18.0 & 45.4 \\
    & 8\,cm
    & 30.1 & 39.5 & 29.9 & 25.2 & 5.6 & 48.0 & 24.5 & 44.4 \\
\cmidrule(lr){1-10}
\multirow{2}{*}{$100 \times 50$}
    & 20\,cm
    & 24.0 & 29.9 & 21.7 & 14.7 & 2.1 & 19.7 & 14.3 & 36.5 \\
    & 8\,cm
    & 25.6 & 29.8 & 19.6 & 15.0 & 1.9 & 37.3 & 17.8 & 22.4 \\
\bottomrule
\end{tabular}
\end{table*}

%% file: sec/tables_figures/appendix_qual_visus.tex
\begin{figure}[H]
    \centering

    \setlength{\tabcolsep}{2.5pt}
    \renewcommand{\arraystretch}{1.0}

    \begin{tabular}{
        @{}
        >{\centering\arraybackslash}m{0.242\textwidth}
        >{\centering\arraybackslash}m{0.242\textwidth}
        >{\centering\arraybackslash}m{0.242\textwidth}
        >{\centering\arraybackslash}m{0.242\textwidth}
        @{}
    }
        \small\textbf{Ground Truth} &
        \small\textbf{StreamMapNet} &
        \small\textbf{AerialFusionMapNet} &
        \small\textbf{\shortstack{AerialFusionMapNet\\Camera View}}
        \\[1.2mm]

        \includegraphics[width=\linewidth]{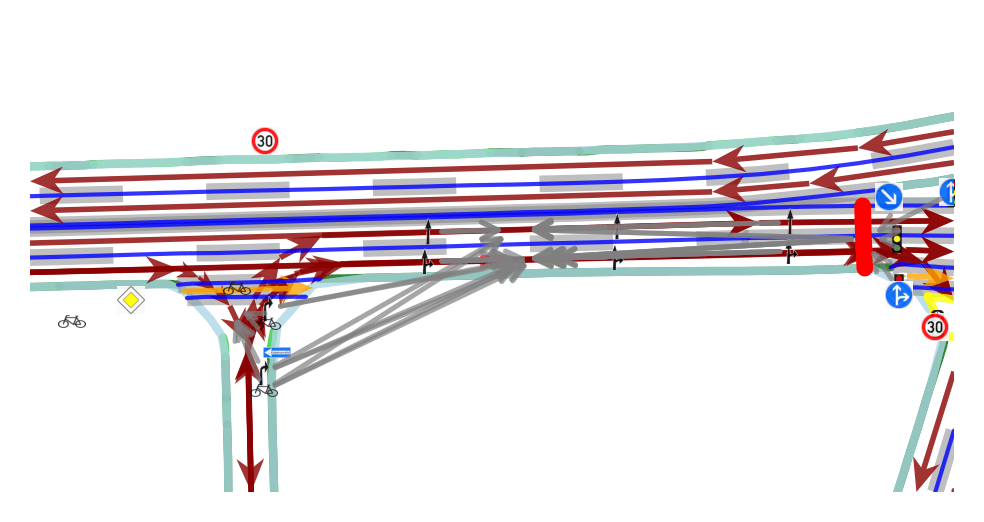} &
        \includegraphics[width=\linewidth]{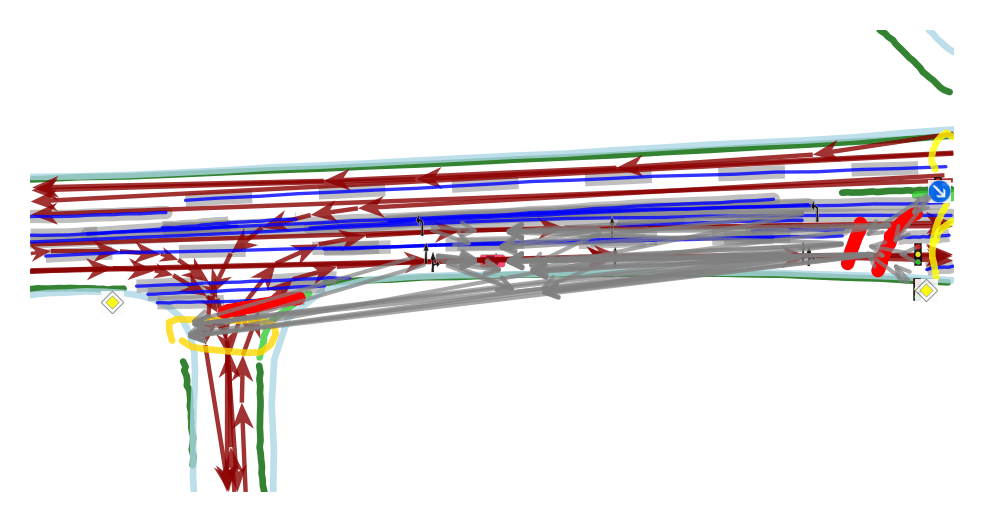} &
        \includegraphics[width=\linewidth]{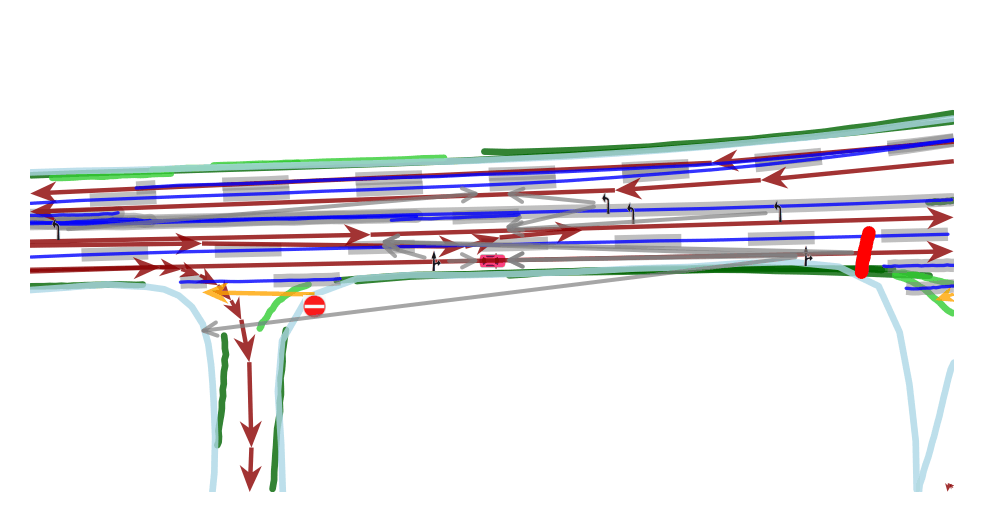} &
        \includegraphics[width=\linewidth]{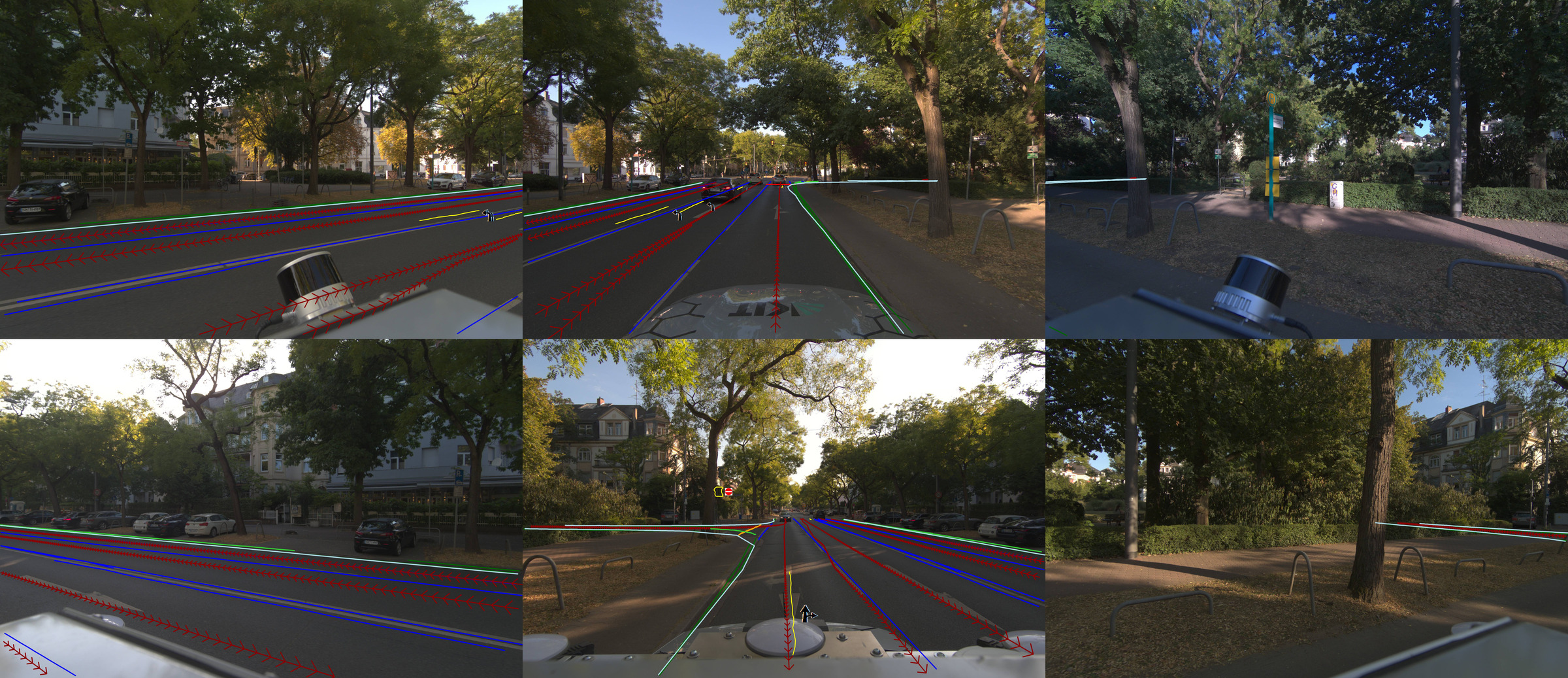}
        \\[1.2mm]
        \includegraphics[width=\linewidth]{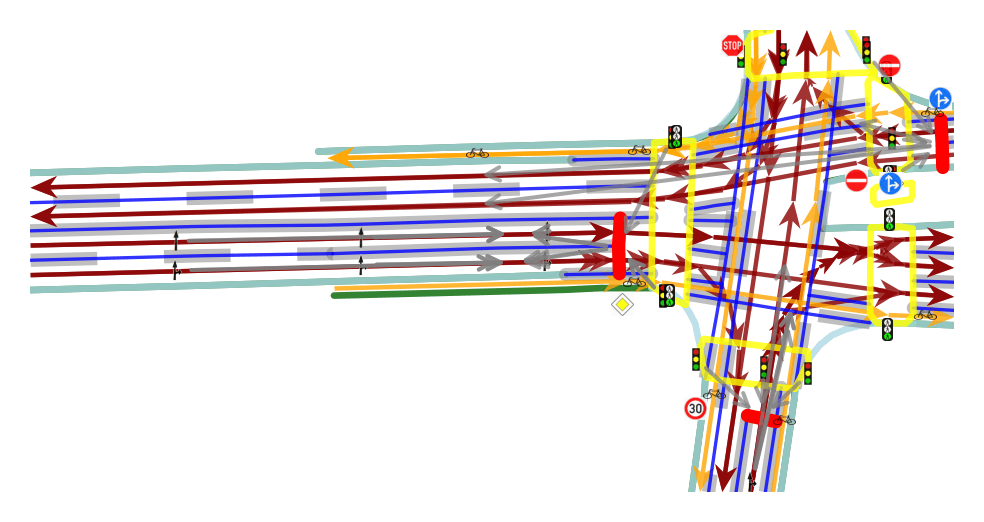} &
        \includegraphics[width=\linewidth]{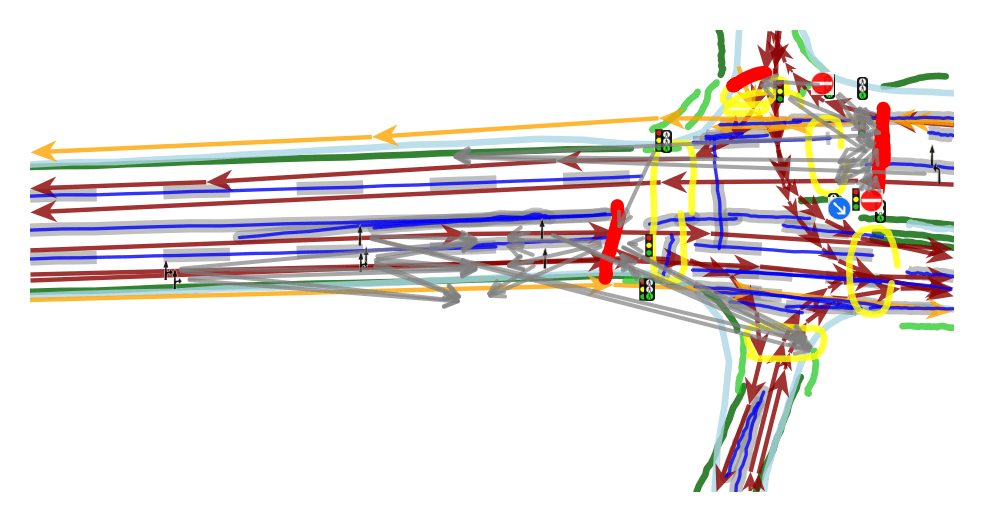} &
        \includegraphics[width=\linewidth]{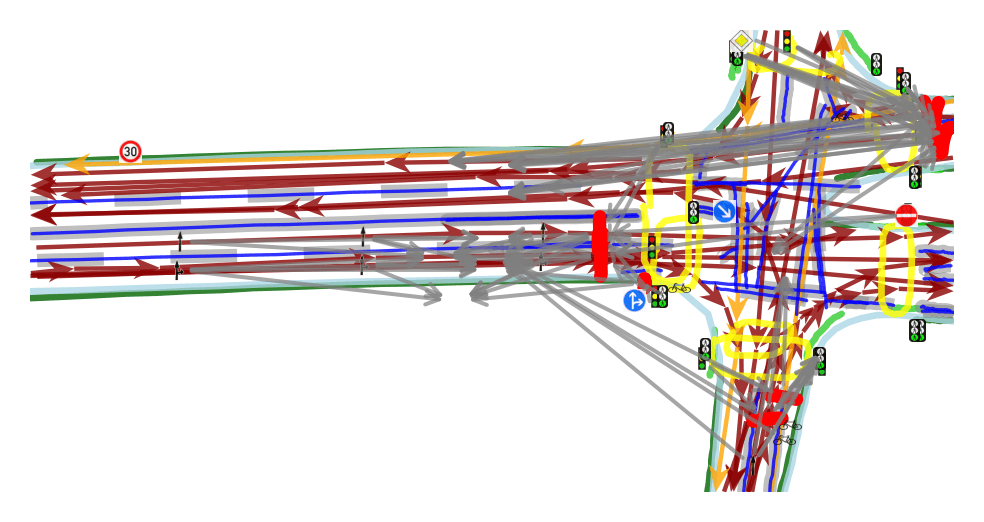} &
        \includegraphics[width=\linewidth]{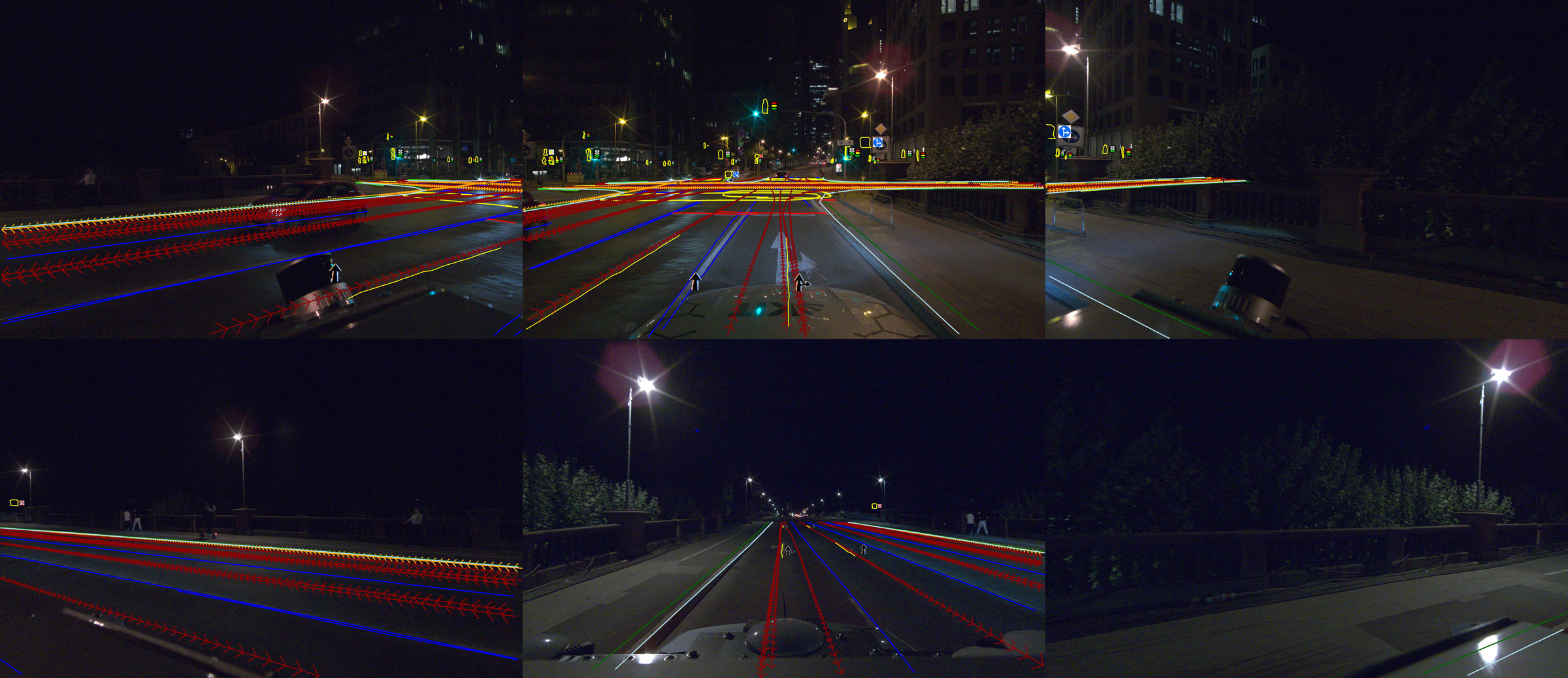}
        \\[1.2mm]
        \includegraphics[width=\linewidth]{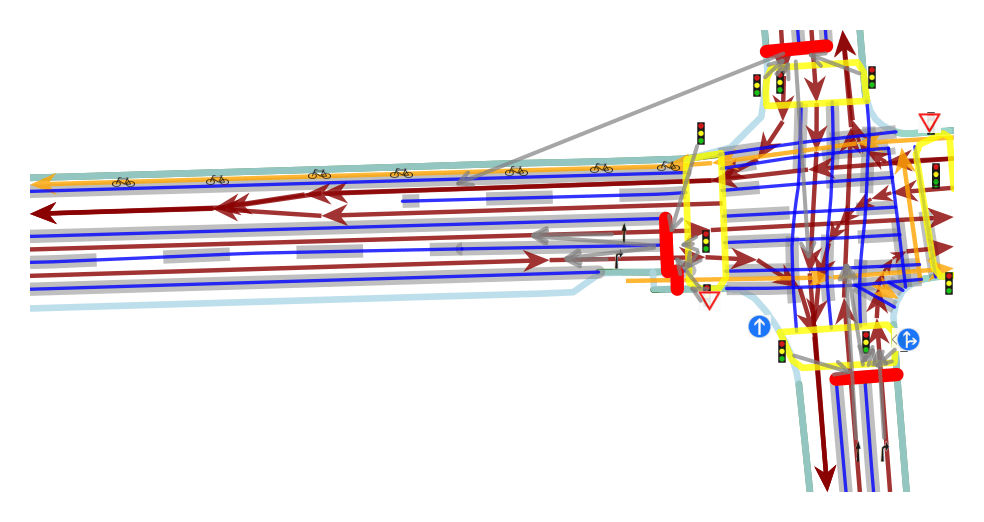} &
        \includegraphics[width=\linewidth]{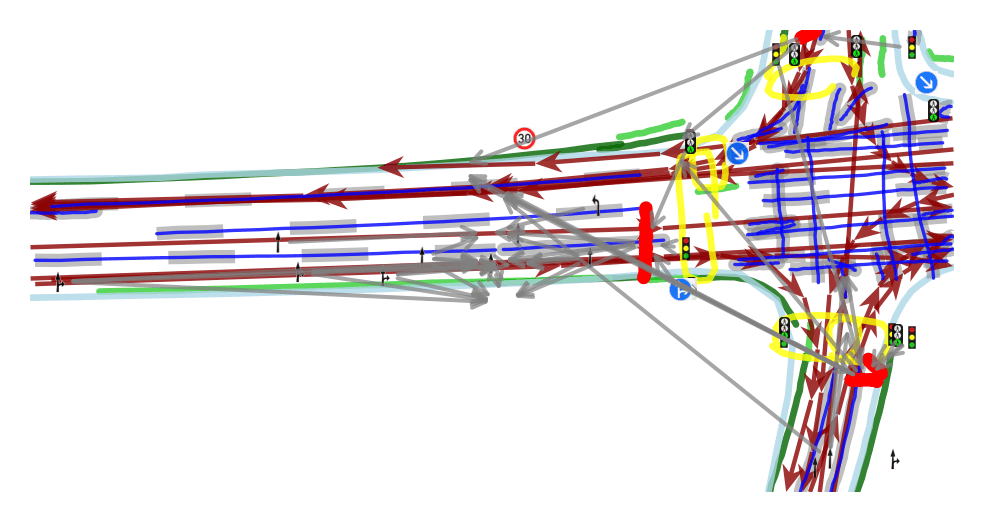} &
        \includegraphics[width=\linewidth]{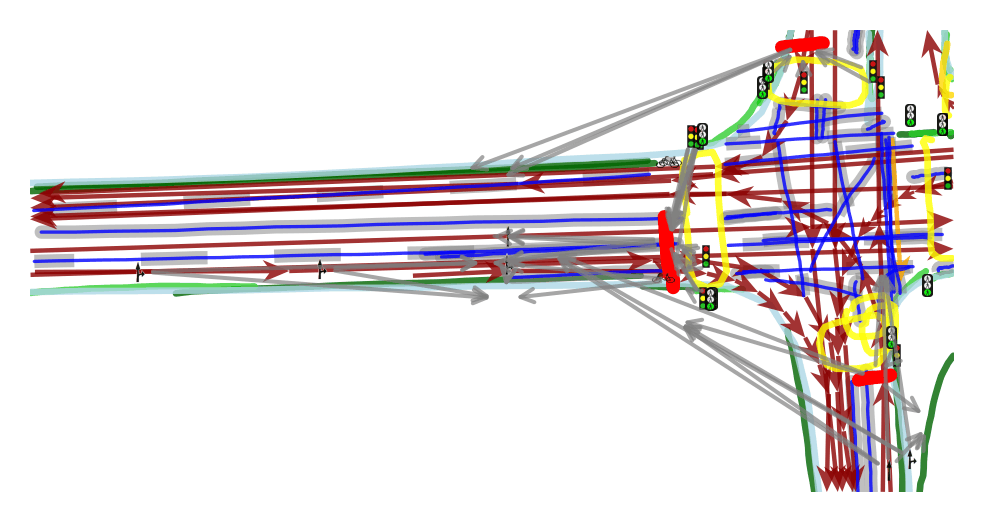} &
        \includegraphics[width=\linewidth]{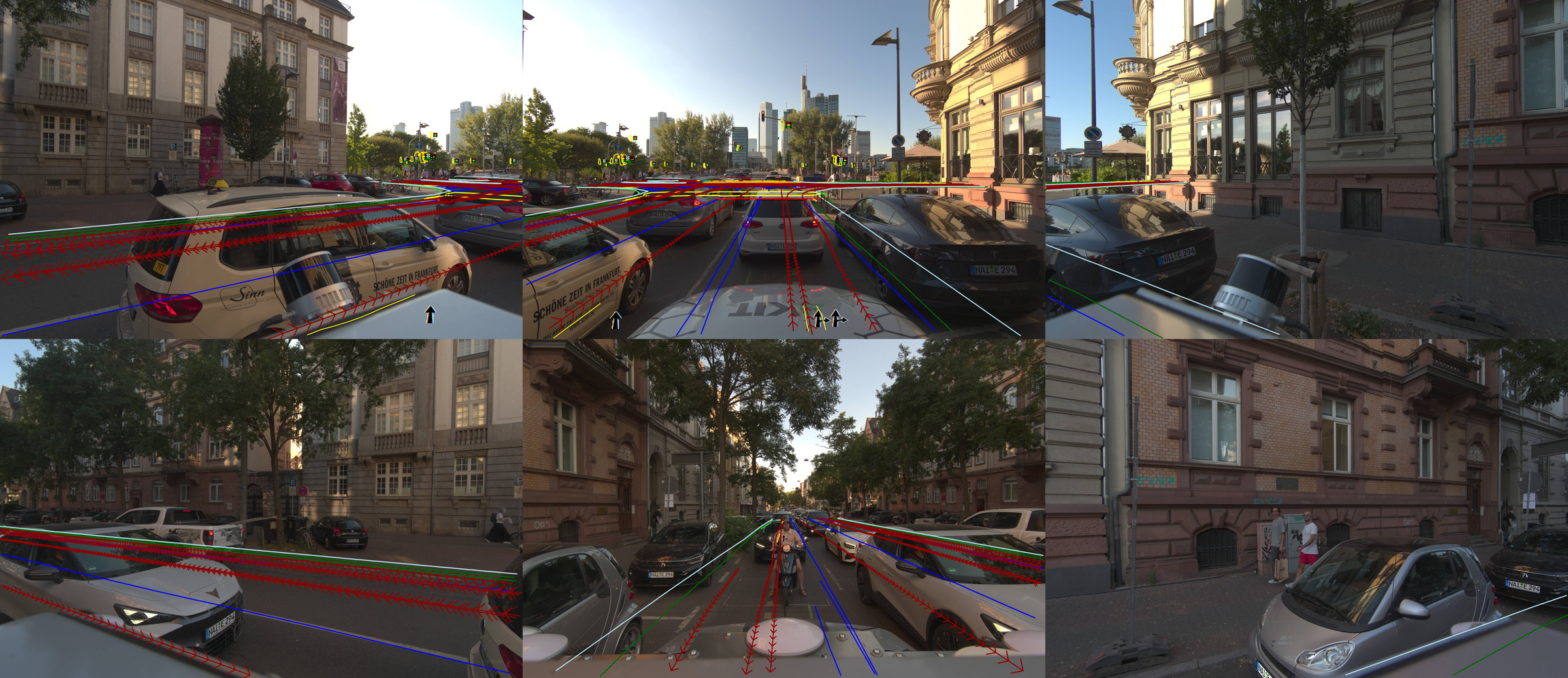}
        \\[1.2mm]
        \includegraphics[width=\linewidth]{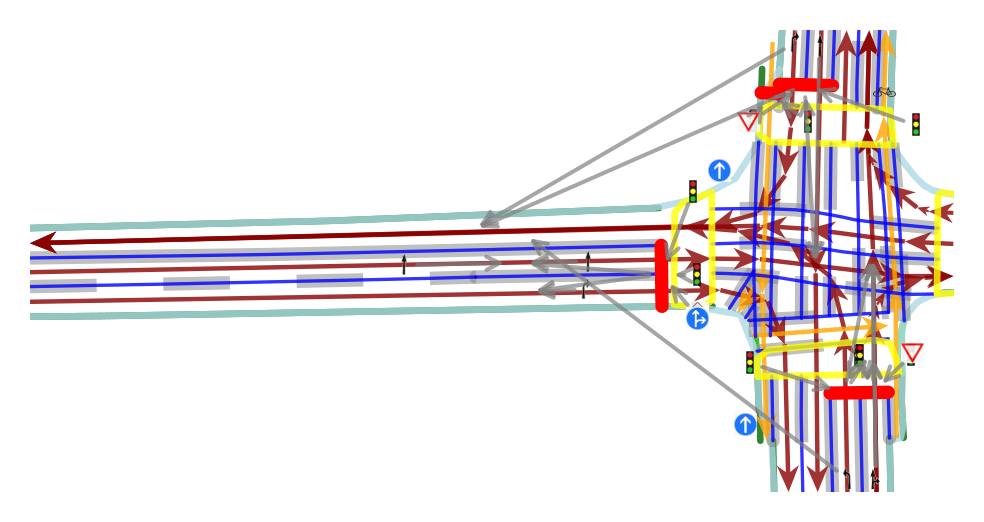} &
        \includegraphics[width=\linewidth]{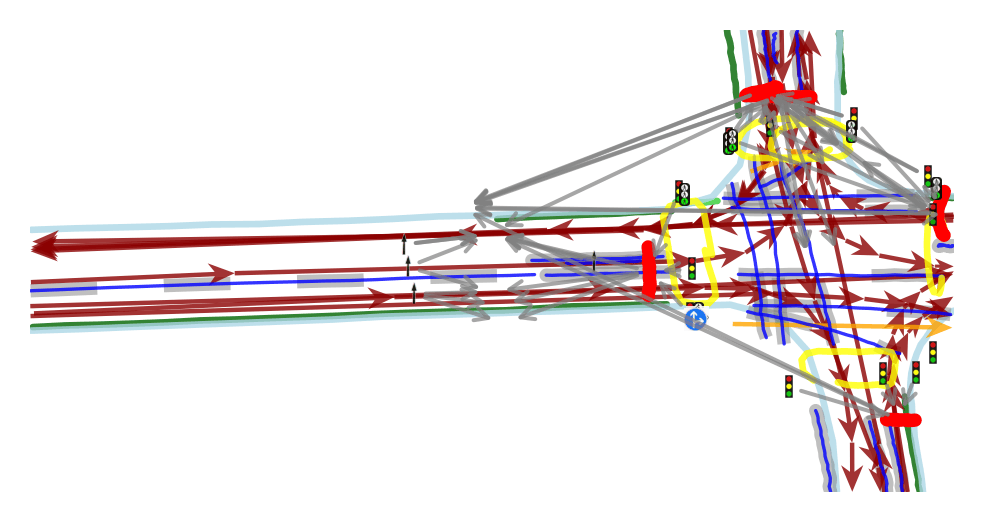} &
        \includegraphics[width=\linewidth]{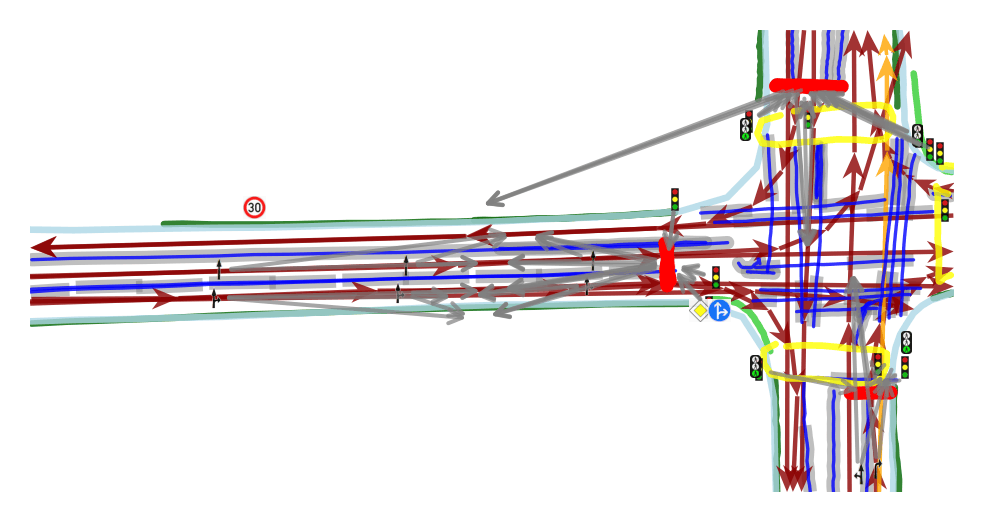} &
        \includegraphics[width=\linewidth]{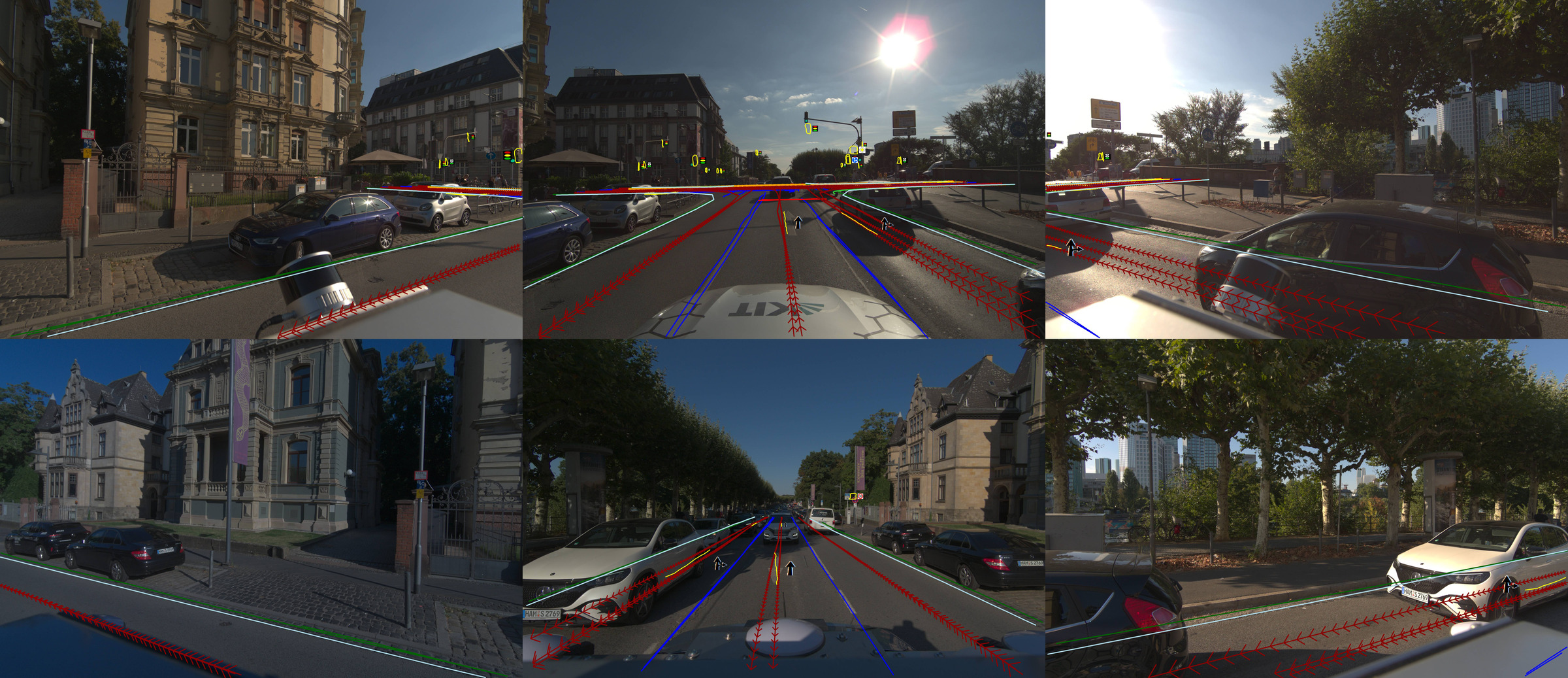}
        \\[1.2mm]

    \end{tabular}

    \vspace{-1mm}

    \caption{
        Qualitative comparison of online HD map construction on KITScenes Aerial.
        Predictions from the onboard-only StreamMapNet baseline are compared
        with the aerial-fused AerialFusionMapNet.
        Camera views show the corresponding ground-level context
        with AerialFusionMapNet detections.
    }
    \label{fig:appendix_qualitative_mapping}
\end{figure}

%% file: sec/tables_figures/stitched_figures.tex
\begin{figure*}
    \centering
    \includegraphics[width=0.95\linewidth]{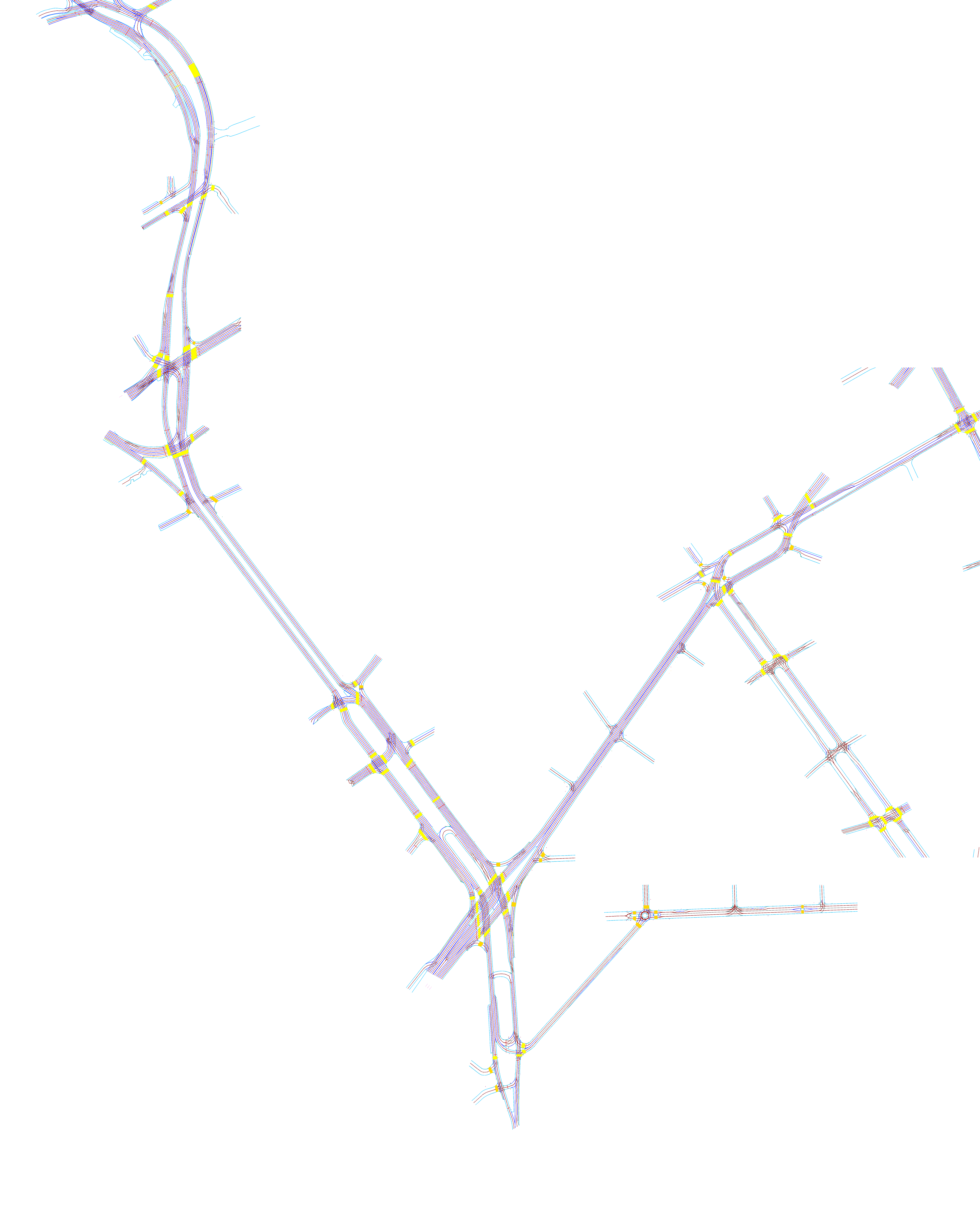}
    \caption{Frankfurt KITScenes rasterized HD Maps, GT Stitched}
    \label{fig:stitched_large_scale_gt}
\end{figure*}

\begin{figure*}
    \centering
    \includegraphics[width=0.95\linewidth]{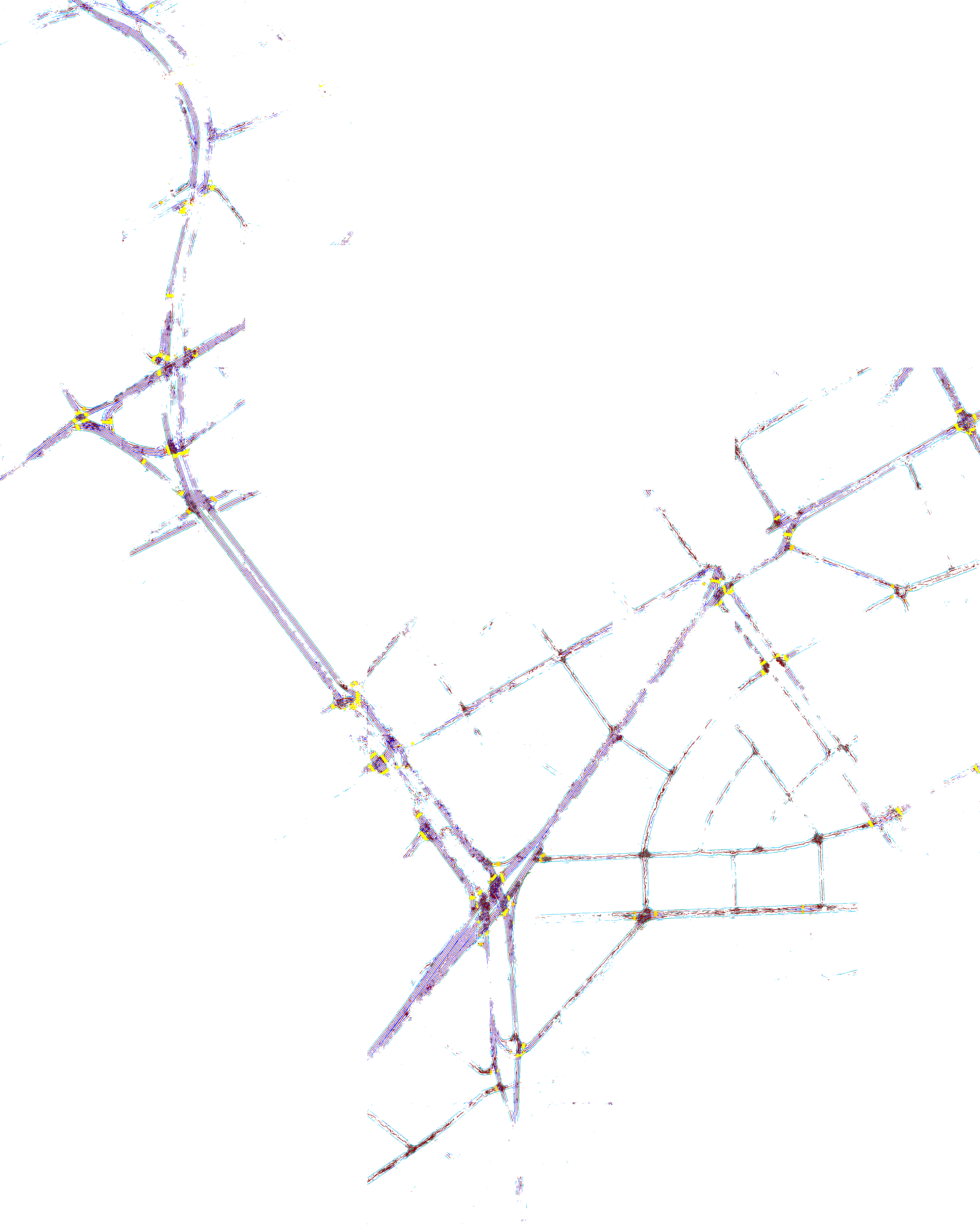}
    \caption{Frankfurt Validation DINOv3 Features + DPT Head Predictions Stitched}
    \label{fig:stitched_large_scale_pred}
\end{figure*}

\begin{figure*}
    \centering
    \includegraphics[width=0.75\linewidth]{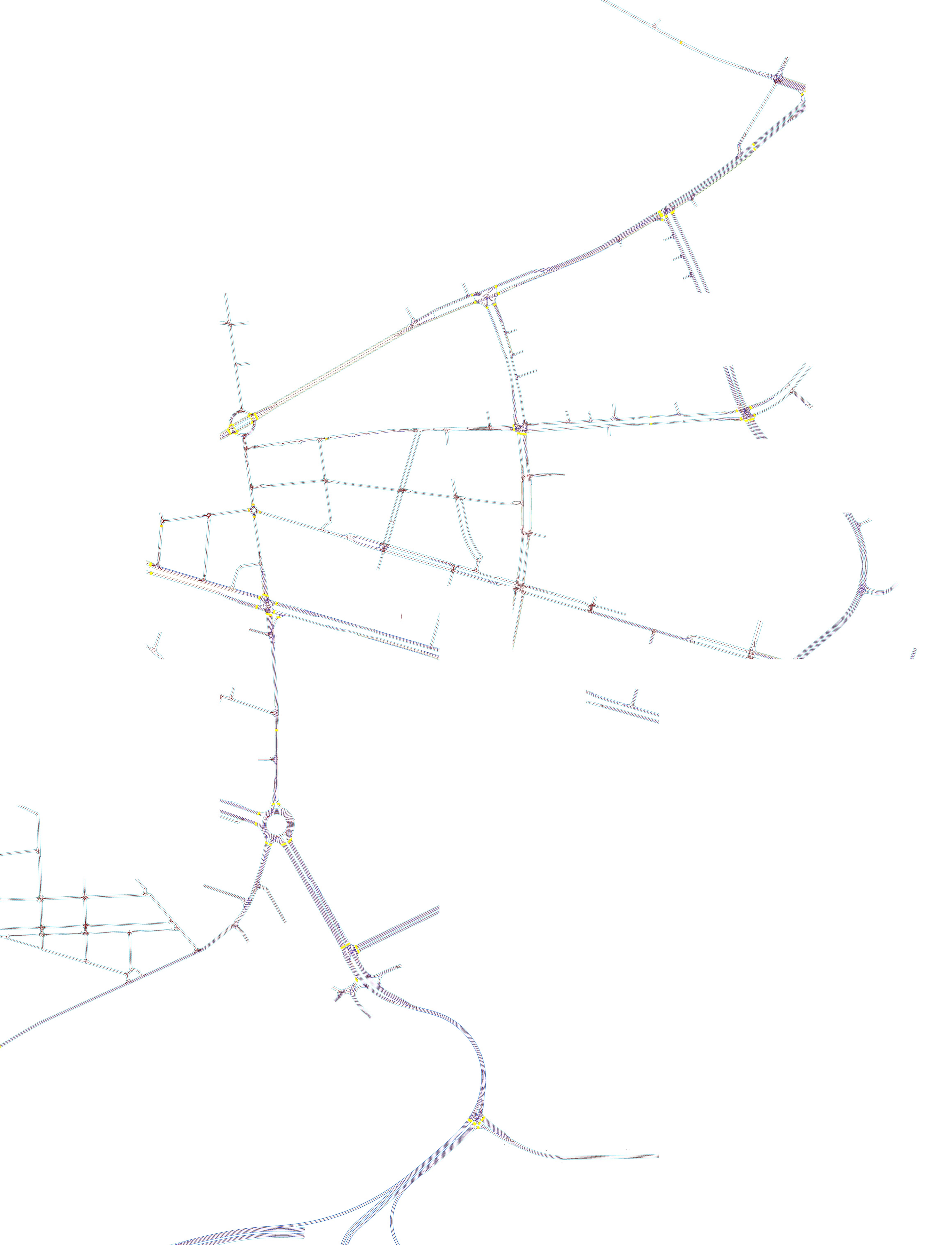}
    \caption{Karlsruhe KITScenes rasterized HD Maps, GT Stitched}
    \label{fig:stitched_large_scale_gt2}
\end{figure*}

\begin{figure*}
    \centering
    \includegraphics[width=0.75\linewidth]{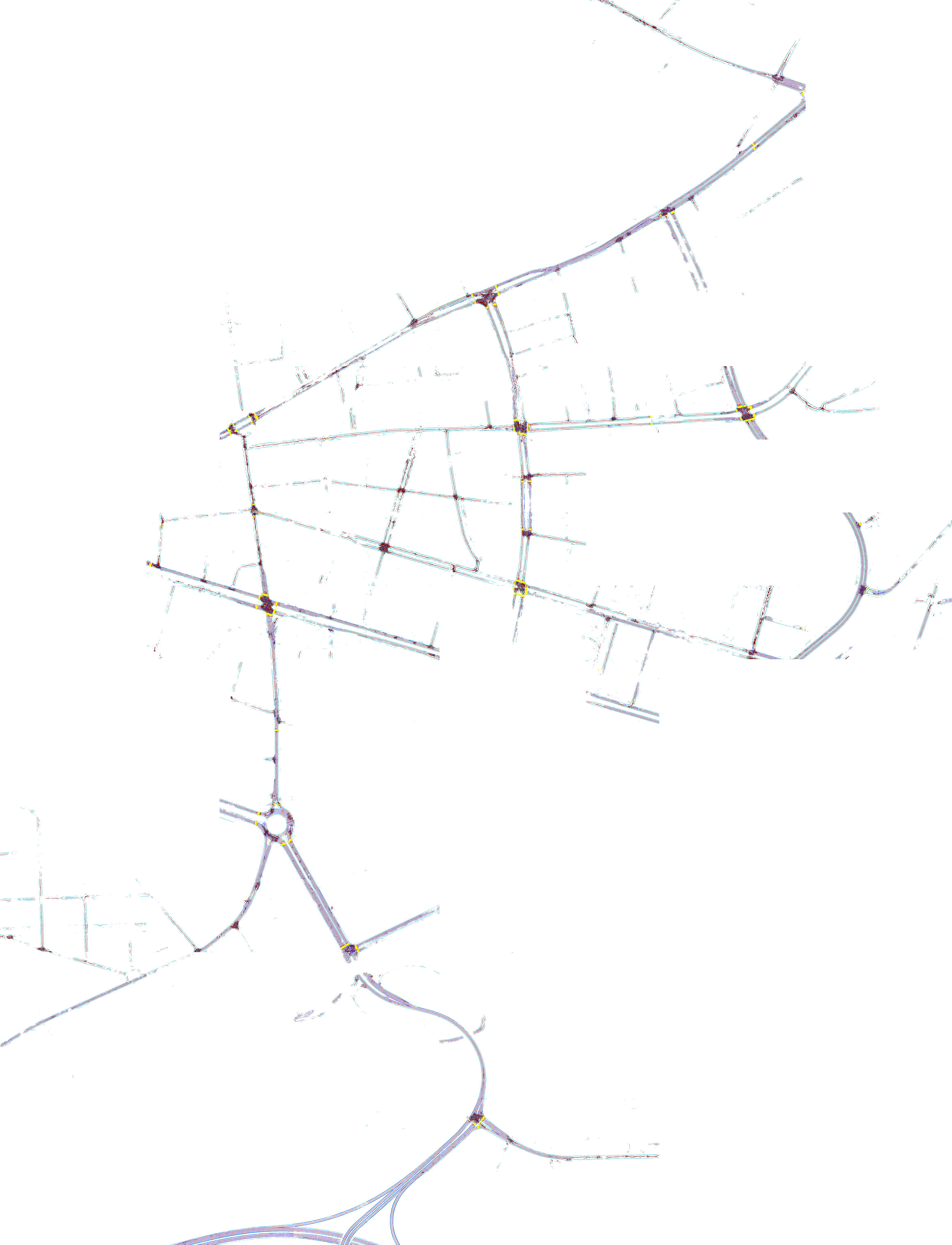}
    \caption{Karlsruhe Validation DINOv3 Features + DPT Head Predictions Stitched}
    \label{fig:stitched_large_scale_pred2}
\end{figure*}